\documentclass[10pt,twocolumn]{article}
\usepackage[utf8]{inputenc}
\usepackage[T1]{fontenc}
\usepackage{times}
\usepackage{microtype}
\usepackage{graphicx}
\graphicspath{{figures/}}
\usepackage{amsmath,amssymb}
\usepackage{booktabs}
\usepackage{hyperref}
\usepackage{cleveref}
\usepackage{xcolor}
\usepackage{algorithm}
\usepackage{algpseudocode}
\usepackage{multirow}
\usepackage{caption}
\usepackage{subcaption}
\usepackage{tikz}
\usepackage{pgfplots}
\pgfplotsset{compat=1.17}
\usetikzlibrary{shapes,arrows,positioning,fit,backgrounds,calc}
\usepackage[margin=1in]{geometry}
\usepackage{adjustbox}
\usepackage{makecell}
\usepackage{enumitem}
\usepackage{tabularx}
\usepackage{url}
\usepackage{float}
\usepackage{etoolbox}
\usepackage{cuted}

\hypersetup{
    colorlinks=true,
    linkcolor=blue!70!black,
    citecolor=green!50!black,
    urlcolor=blue!70!black,
}

\newcommand{\carlaair}{\textsc{Carla-Air}}

\title{\carlaair{}: Fly Drones Inside a CARLA World\\
A Unified Infrastructure for Air-Ground Embodied Intelligence}

\author{
    Tianle Zeng$^{1}$ \quad
    Yanci Wen$^{1}$ \quad
    Hong Zhang$^{1,\dagger}$
}

\date{}

\makeatletter
\apptocmd{\maketitle}{%
    {%
        
        \footnotetext{%

            $^{\dagger}$Corresponding author (hzhang@sustech.edu.cn).
            
            $^{1}$Shenzhen Key Laboratory of Robotics and Computer Vision,
            Southern University of Science and Technology.

            For questions and technical inquiries, please contact
            \texttt{louiszeng16@163.com}.%
        }%
    }%
    \begin{strip}
        \centering
        \vspace*{-25mm}
        \includegraphics[width=\textwidth]{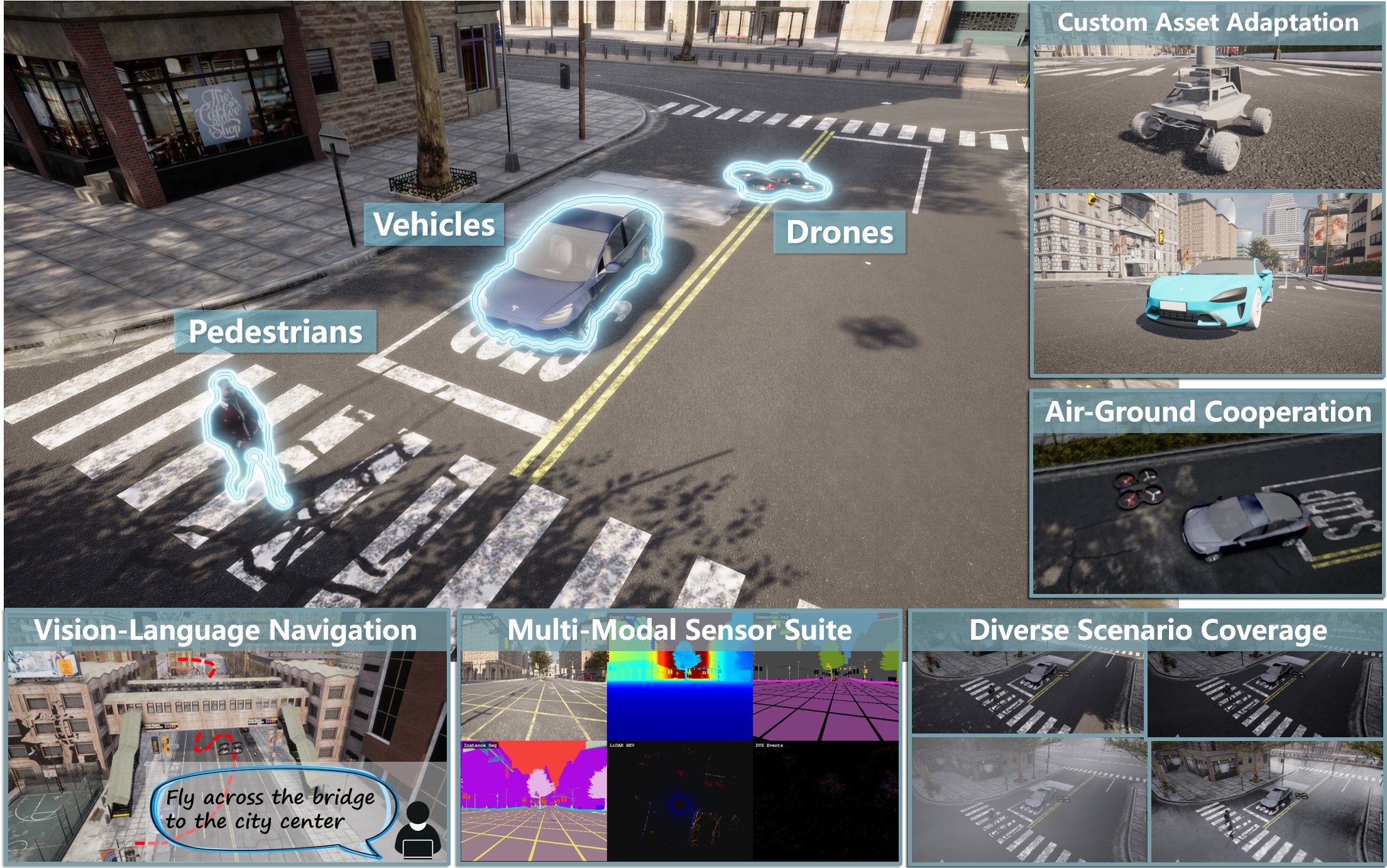}
        \captionof{figure}{Overview of \textbf{CARLA-Air}, a unified simulation
        infrastructure for air-ground embodied intelligence. The examples shown
        here illustrate representative capabilities of the platform, including
        unified air-ground simulation, multi-modal sensing, embodied navigation,
        asset adaptation, and diverse urban scenarios within a single physically
        coherent environment.}
        \label{fig:teaser}
        \vspace*{-2mm}
    \end{strip}
    \setcounter{figure}{1}
    \thispagestyle{empty}
}{}{}
\makeatother

\begin{document}
\sloppy
\raggedbottom
\hbadness=10000
\vbadness=10000

\maketitle


\begin{abstract}
The convergence of low-altitude economies, embodied intelligence, and
air-ground cooperative systems creates a growing need for simulation
infrastructure capable of jointly modeling aerial and ground agents within a
single physically coherent environment.
Existing open-source platforms remain domain-segregated: urban driving
simulators provide rich traffic populations but no aerial dynamics, while
multirotor simulators offer physics-accurate flight but lack realistic ground
scenes.
Bridge-based co-simulation can connect heterogeneous backends, yet introduces
synchronization overhead and cannot guarantee the strict spatial-temporal
consistency required by modern perception and learning pipelines.

We present \textbf{CARLA-Air}, an open-source infrastructure that unifies
high-fidelity urban driving and physics-accurate multirotor flight within a
single Unreal Engine process, providing a practical simulation foundation for
air-ground embodied intelligence research.
The platform preserves both CARLA and AirSim native Python APIs and
ROS\,2 interfaces, enabling zero-modification reuse of existing codebases.
Within a shared physics tick and rendering pipeline, CARLA-Air delivers
photorealistic urban and natural environments populated with rule-compliant
traffic flow, socially-aware pedestrians, and aerodynamically consistent UAV
dynamics, while synchronously capturing up to 18 sensor modalities---including
RGB, depth, semantic segmentation, LiDAR, radar, IMU, GNSS, and
barometry---across all aerial and ground platforms at each simulation tick.
Building on this foundation, the platform provides out-of-the-box support for
representative air-ground embodied intelligence workloads, spanning air-ground
cooperation, embodied navigation and vision-language action, multi-modal
perception and dataset construction, and reinforcement-learning-based policy
training.
An extensible asset pipeline further allows researchers to integrate custom
robot platforms, UAV configurations, and environment maps into the shared
simulation world.
By inheriting and extending the aerial simulation capabilities of
AirSim---whose upstream development has been archived---CARLA-Air also ensures
that this widely adopted flight stack continues to evolve within a modern,
actively maintained infrastructure.

CARLA-Air is released with both prebuilt binaries and full source code to
support immediate adoption:
\url{https://github.com/louiszengCN/CarlaAir}
\end{abstract}

\section{Introduction}
\label{sec:introduction}

Three converging frontiers are reshaping autonomous systems research.
The \emph{low-altitude economy} demands scalable infrastructure for urban air
mobility, drone logistics, and aerial inspection.
\emph{Embodied intelligence} requires agents that perceive and act in shared
physical environments through vision, language, and control.
\emph{Air-ground cooperative systems} bring these threads together, calling for
heterogeneous robots that operate jointly across aerial and ground domains.
Simulation is essential for advancing all three frontiers, as real-world
deployment is costly, safety-critical, and difficult to scale.
Yet no widely adopted open-source platform provides a unified infrastructure
capable of jointly modeling aerial and ground agents within a single physically
coherent environment.
CARLA-Air is designed to fill this gap.

\paragraph{The simulation landscape and its gap.}
Existing open-source simulators address complementary domains without overlap.
CARLA~\cite{carla}, built on Unreal Engine~4~\cite{ue4}, has become the de
facto standard for urban autonomous driving research, offering photorealistic
environments, rich traffic populations, and a mature Python API.
AirSim~\cite{airsim}, also built on UE4, provides physics-accurate multirotor
simulation with high-frequency dynamics and a comprehensive aerial sensor suite.
The limitation of each platform is precisely the strength of the other: CARLA
lacks aerial agents, while AirSim lacks realistic ground traffic and pedestrian
interactions.
Meanwhile, AirSim's upstream development has been archived by its original
maintainers, leaving a widely adopted flight simulation stack without an active
evolution path.
Other simulators across driving, UAV, and embodied AI domains similarly remain
confined to a single agent modality (see Section~\ref{sec:related} for a
comprehensive survey).
As a result, emerging workloads that span both air and ground
domains---air-ground cooperation, cross-domain embodied navigation, joint
multi-modal data collection, and cooperative reinforcement learning---lack a
shared simulation foundation.

\begin{figure}[t]
\centering
\begin{tikzpicture}
\begin{axis}[
    ybar,
    width=\columnwidth,
    height=4.5cm,
    bar width=7pt,
    xlabel={Number of concurrent sensors},
    ylabel={Per-frame data transfer (ms)},
    xtick={1,4,8,12,16},
    ymin=0, ymax=26,
    ytick={0,5,10,15,20,25},
    legend style={
        at={(0.03,0.97)}, anchor=north west,
        font=\footnotesize,
        draw=gray!40,
        fill=white,
        inner sep=3pt,
        row sep=1pt,
    },
    tick label style={font=\footnotesize},
    label style={font=\small},
    grid=major,
    grid style={dotted, gray!30},
    every axis plot/.append style={thick},
    enlarge x limits=0.12,
]
\addplot[
    fill=red!25!white,
    draw=red!60!black,
] coordinates {
    (1,  1.3)
    (4,  5.2)
    (8, 10.6)
    (12,15.8)
    (16,20.4)
};
\addplot[
    fill=blue!30!white,
    draw=blue!60!black,
] coordinates {
    (1,  0.32)
    (4,  0.34)
    (8,  0.38)
    (12, 0.41)
    (16, 0.45)
};
\legend{Bridge co-sim~\cite{transimhub}, CARLA-Air (ours)}
\end{axis}
\end{tikzpicture}
\caption{%
  Per-frame inter-process data transfer time as a function of concurrent sensor
  count.
  Bridge-based co-simulation~\cite{transimhub} exhibits near-linear growth with
  sensor count due to cross-process serialization, while CARLA-Air remains
  effectively constant ($<0.5$\,ms) owing to its single-process architecture.%
}
\label{fig:ipc_overhead}
\end{figure}
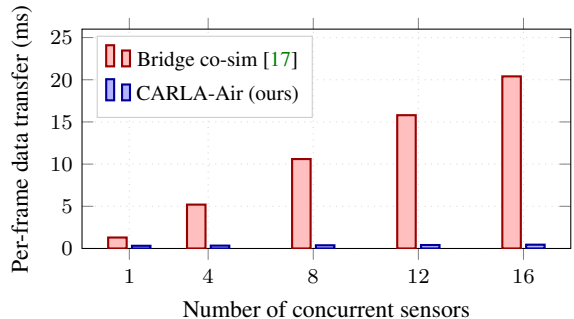

\paragraph{Why not bridge-based co-simulation?}
A common workaround connects heterogeneous simulators through bridge-based
co-simulation, typically via ROS\,2~\cite{ros2} or custom message-passing
interfaces.
While functionally viable, such approaches introduce inter-process
synchronization complexity, communication overhead, and duplicated rendering
pipelines.
More critically, independent simulation processes cannot guarantee strict
spatial-temporal consistency across sensor streams---a requirement for
perception, learning, and evaluation in embodied intelligence systems.
Fig.~\ref{fig:ipc_overhead} quantifies the per-frame inter-process overhead
contrast between bridge-based co-simulation and the single-process design
adopted by CARLA-Air.


\paragraph{CARLA-Air: a unified infrastructure for air-ground embodied intelligence.}
We present \textbf{CARLA-Air}, an open-source platform that integrates CARLA
and AirSim within a single Unreal Engine process, purpose-built as a practical
simulation foundation for air-ground embodied intelligence research.
By inheriting and extending AirSim's aerial simulation capabilities within a
modern, actively maintained infrastructure, CARLA-Air also provides a
sustainable evolution path for the large body of existing AirSim-based research.
Key capabilities of the platform include:

\begin{enumerate}[leftmargin=*, label=(\roman*)]

\item \textbf{Single-process air-ground integration.}
CARLA-Air resolves a fundamental engine-level conflict---UE4 permits only one
active game mode per world---through a composition-based design that inherits
all ground simulation subsystems from CARLA while spawning AirSim's aerial
flight actor as a regular world entity.
This yields a shared physics tick, a shared rendering pipeline, and strict
spatial-temporal consistency across all sensor viewpoints.

\item \textbf{Full API compatibility and zero-modification code migration.}
Both CARLA and AirSim native Python APIs and ROS\,2 interfaces are fully
preserved, allowing existing research codebases to run on CARLA-Air without
modification.

\item \textbf{Photorealistic, physically coherent simulation world.}
The platform delivers rich urban and natural environments populated with
rule-compliant traffic flow, socially-aware pedestrians, and aerodynamically
consistent multirotor dynamics, with synchronized capture of up to 18 sensor
modalities across all aerial and ground platforms at each simulation tick.

\item \textbf{Extensible asset pipeline.}
Researchers can import custom robot platforms, UAV configurations, vehicles,
and environment maps into the shared simulation world, enabling flexible
construction of diverse air-ground interaction scenarios.

\end{enumerate}

\noindent Building on these capabilities, CARLA-Air provides out-of-the-box
support for representative air-ground embodied intelligence workloads across
four research directions:

\begin{enumerate}[leftmargin=*, label=(\alph*)]

\item \textbf{Air-ground cooperation}---heterogeneous aerial and ground agents
coordinate within a shared environment for tasks such as cooperative
surveillance, escort, and search-and-rescue.

\item \textbf{Embodied navigation and vision-language action}---agents navigate
and act grounded in visual and linguistic input, leveraging both aerial
overview and ground-level detail.

\item \textbf{Multi-modal perception and dataset construction}---synchronized
aerial-ground sensor streams are collected at scale to build paired datasets
for cross-view perception, 3D reconstruction, and scene understanding.

\item \textbf{Reinforcement-learning-based policy training}---agents learn
cooperative or individual policies through closed-loop interaction in
physically consistent air-ground environments.

\end{enumerate}

\noindent As a lightweight and practical infrastructure, CARLA-Air lowers the
barrier for developing and evaluating air-ground embodied intelligence systems,
and provides a unified simulation foundation for emerging applications in
low-altitude robotics, cross-domain autonomy, and large-scale embodied AI
research.


\section{Related Work}
\label{sec:related}

Simulation platforms relevant to autonomous systems span autonomous driving,
aerial robotics, joint co-simulation, and embodied AI.
From the perspective of air-ground embodied intelligence, the central question
is not whether a platform supports driving or flight in isolation, but whether
aerial and ground agents can be jointly simulated within a unified, physically
coherent, and practically usable environment.
As illustrated in Fig.~\ref{fig:platform_positioning} and
Table~\ref{tab:platform_comparison}, existing open-source platforms largely
remain separated by domain focus, and none simultaneously provides realistic
urban traffic, socially-aware pedestrians, physics-based multirotor flight,
preserved native APIs, and single-process execution in one shared simulation
world.

\subsection{Autonomous Driving Simulators}
\label{sec:related:driving}

Autonomous driving simulators provide strong support for realistic urban
scenes, traffic agents, and ground-vehicle perception.
CARLA~\cite{carla}, built on Unreal Engine~\cite{ue4}, has become the de facto
open-source platform for urban driving research due to its photorealistic
environments, rich actor library, and mature Python API.
LGSVL~\cite{lgsvl} offers full-stack integration with Autoware and Apollo on
the Unity engine.
SUMO~\cite{sumo} provides lightweight microscopic traffic flow modeling.
MetaDrive~\cite{metadrive} enables procedural environment generation for
generalizable RL, and VISTA~\cite{vista} supports data-driven sensor-view
synthesis for autonomous vehicles.
These platforms collectively cover a broad range of ground-autonomy research
needs, but none natively supports physics-based UAV flight, leaving air-ground
cooperative workloads outside their scope.

\subsection{Aerial Vehicle Simulators}
\label{sec:related:uav}

Aerial simulators provide the complementary capability: accurate multirotor
dynamics, onboard aerial sensing, and UAV-oriented control interfaces.
AirSim~\cite{airsim} remains one of the most widely adopted open-source UAV
simulators, offering physics-accurate multirotor flight and a comprehensive
sensor suite on Unreal Engine, though its upstream development has since been
archived.
Flightmare~\cite{flightmare} combines Unity-based photorealistic rendering
with highly parallel dynamics for fast RL training.
FlightGoggles~\cite{flightgoggles} provides photogrammetry-based environments
for perception-driven aerial robotics.
Gazebo~\cite{gazebo}, together with MAV-specific packages such as
RotorS~\cite{rotorsim}, offers a mature ROS-integrated simulation stack for
multi-rotor control and state estimation.
OmniDrones~\cite{omnidrones} and
gym-pybullet-drones~\cite{pybulletdrones} target scalable, GPU-accelerated or
lightweight RL-oriented multi-agent UAV training.
While these systems are well suited to aerial autonomy in isolation, they
generally lack realistic urban traffic populations, pedestrian interactions, and
richly populated ground environments, limiting their use as infrastructure for
air-ground cooperation or cross-domain data collection.

\subsection{Joint and Co-Simulation Platforms}
\label{sec:related:joint}

The most relevant prior efforts attempt to combine aerial and ground simulation
through co-simulation.
TranSimHub~\cite{transimhub} connects CARLA with SUMO and aerial agents via a
multi-process architecture supporting synchronized multi-view rendering.
Other representative approaches include ROS-based pairings of AirSim with
Gazebo~\cite{airsim,gazebo,ros2}.
These systems demonstrate that heterogeneous simulation backends can be
functionally connected, but their integration typically depends on bridges, RPC
layers, or message-passing middleware across independent processes.
As summarized in Table~\ref{tab:related_arch}, such designs do not preserve a
single rendering pipeline, do not provide strict shared-tick execution, and
often require adapting existing code to new interfaces.
By contrast, CARLA-Air integrates both simulation backends within a single
Unreal Engine process, preserving both native APIs while maintaining a shared
world state, shared renderer, and synchronized sensing---a system-level
distinction detailed in Section~\ref{sec:architecture}.

\subsection{Embodied AI and Robot Learning Platforms}
\label{sec:related:embodied}

Embodied AI platforms prioritize a different design objective: scalable policy
training rather than realistic urban air-ground infrastructure.
Isaac Lab~\cite{isaaclab} and Isaac Gym~\cite{isaacgym} emphasize massively
parallel GPU-accelerated reinforcement learning for locomotion and
manipulation.
Habitat~\cite{habitat} and SAPIEN~\cite{sapien} target indoor navigation and
articulated object interaction, while RoboSuite~\cite{robosuite} focuses on
tabletop manipulation benchmarks.
These platforms are valuable for embodied intelligence research, but they do
not provide the urban traffic realism, socially-aware pedestrian populations,
or integrated aerial-ground simulation required by low-altitude cooperative
robotics.
In this sense, they address complementary research needs and are not direct
substitutes for CARLA-Air.

\subsection{Summary}
\label{sec:related:summary}

\begin{figure}[!htbp]
\centering
\adjustbox{max width=\columnwidth, max height=0.32\textheight}{
\begin{tikzpicture}[
  font=\scriptsize,
  every node/.style={align=center},
  platform/.style={
    draw, rounded corners=2pt, fill=white,
    minimum width=1.5cm, minimum height=0.48cm,
    font=\tiny\sffamily, semithick
  },
  ours/.style={
    draw=black!80, rounded corners=2pt,
    fill=black!10, minimum width=1.6cm, minimum height=0.52cm,
    font=\tiny\sffamily\bfseries, semithick, line width=1pt
  },
  axis/.style={-latex, semithick, black!70},
  gridline/.style={dashed, black!15, thin}
]

\draw[axis] (0,0) -- (7.8,0)
  node[right, font=\footnotesize] {Agent Domain Breadth};
\draw[axis] (0,0) -- (0,6.2)
  node[above, font=\footnotesize] {Simulation Fidelity};

\node[font=\scriptsize, text=black!55] at (1.8,-0.35) {Single domain};
\node[font=\scriptsize, text=black!55] at (5.8,-0.35) {Multi-domain};
\node[font=\scriptsize, text=black!55, rotate=90] at (-0.42,1.5) {Lightweight};
\node[font=\scriptsize, text=black!55, rotate=90] at (-0.42,4.6) {High-fidelity};

\draw[gridline] (0,3.0) -- (7.5,3.0);
\draw[gridline] (3.8,0) -- (3.8,6.0);

\node[platform, fill=blue!8, draw=blue!50] at (1.7,5.0)
  {CARLA~\cite{carla}\\[-1pt]{\tiny Ground}};
\node[platform, fill=cyan!8, draw=cyan!50] at (2.4,4.3)
  {AirSim~\cite{airsim}\\[-1pt]{\tiny Aerial}};
\node[platform, fill=cyan!5, draw=cyan!30] at (1.5,3.7)
  {FlightGoggles\\[-1pt]{\tiny\cite{flightgoggles}}};
\node[platform, fill=orange!8, draw=orange!50] at (2.7,5.5)
  {Isaac Lab\\[-1pt]{\tiny\cite{isaaclab}}};
\node[platform, fill=green!6, draw=green!40] at (1.9,2.5)
  {Habitat\\[-1pt]{\tiny\cite{habitat}}};
\node[platform, fill=gray!8, draw=gray!40] at (4.4,1.4)
  {SUMO\\[-1pt]{\tiny\cite{sumo}}};
\node[platform, fill=purple!6, draw=purple!40] at (5.5,3.7)
  {TranSimHub\\[-1pt]{\tiny\cite{transimhub}}};
\node[platform, fill=cyan!6, draw=cyan!40] at (4.2,4.3)
  {OmniDrones\\[-1pt]{\tiny\cite{omnidrones}}};
\node[platform, fill=blue!5, draw=blue!30] at (5.0,2.8)
  {MetaDrive\\[-1pt]{\tiny\cite{metadrive}}};

\node[ours] (ca) at (6.5,5.2)
  {CARLA-Air\\[-1pt]{\tiny (Ours)}};

\draw[-latex, dashed, thin, black!40]
  (2.5,5.0) to[out=10, in=185] (ca.west);
\draw[-latex, dashed, thin, black!40]
  (3.2,4.3) to[out=5, in=195] (ca.west);

\node[font=\small] at (6.5,5.7) {$\star$};

\end{tikzpicture}
}
\caption{%
  Platform positioning along simulation fidelity and agent domain breadth.
  CARLA-Air ($\star$) occupies the high-fidelity, multi-domain quadrant
  without inter-process bridging.
  Dashed arrows indicate subsumption of upstream capabilities.%
}
\label{fig:platform_positioning}
\end{figure}
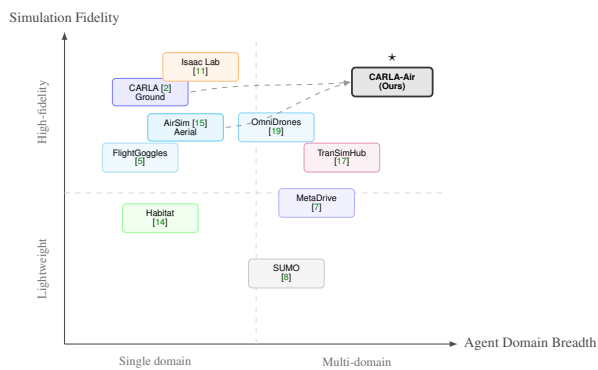

Fig.~\ref{fig:platform_positioning} positions CARLA-Air and representative
platforms along two principal design axes: simulation fidelity and agent domain
breadth.
Driving simulators provide realistic urban ground environments without aerial
dynamics; UAV simulators provide flight realism without populated ground
worlds; joint simulators generally rely on multi-process bridging that
sacrifices interface compatibility or synchronization fidelity; and embodied AI
platforms focus on scalable learning rather than air-ground infrastructure.
CARLA-Air is designed to sit at the intersection of these domains, combining
realistic urban traffic, socially-aware pedestrians, physics-based multirotor
flight, preserved native APIs, and single-process execution within one unified
simulation environment.
Table~\ref{tab:platform_comparison} provides a detailed feature-level comparison
across all platforms discussed above.


\begin{table*}[!htbp]
\centering
\caption{%
  Consolidated comparison of representative open-source simulation platforms
  for air-ground embodied intelligence research.
  \checkmark\,=\,supported;\; $\sim$\,=\,partial or constrained;\;
  \texttimes\,=\,not supported;\; ---\,=\,not applicable.%
}
\label{tab:platform_comparison}
\setlength{\tabcolsep}{2.8pt}
\renewcommand{\arraystretch}{1.06}
\scriptsize
\begin{tabular}{@{}lccccccccccc@{}}
\toprule
\textbf{Platform} &
\textbf{\makecell{Urban\\Traffic}} &
\textbf{\makecell{Pedes-\\trians}} &
\textbf{\makecell{UAV\\Flight}} &
\textbf{\makecell{Single\\Process}} &
\textbf{\makecell{Shared\\Renderer}} &
\textbf{\makecell{Native\\APIs}} &
\textbf{\makecell{Joint\\Sensors}} &
\textbf{\makecell{Prebuilt\\Binary}} &
\textbf{\makecell{Test\\Suite}} &
\textbf{\makecell{Custom\\Assets}} &
\textbf{\makecell{Open\\Source}} \\
\midrule
\multicolumn{12}{@{}l}{\textit{Autonomous Driving}} \\
CARLA~\cite{carla}
  & \checkmark & \checkmark & \texttimes & \checkmark & \checkmark
  & \checkmark & \texttimes & \checkmark & \texttimes & \checkmark & \checkmark \\
LGSVL~\cite{lgsvl}
  & \checkmark & \checkmark & \texttimes & \checkmark & \checkmark
  & \checkmark & \texttimes & \checkmark & \texttimes & $\sim$ & \checkmark \\
SUMO~\cite{sumo}
  & \checkmark & \checkmark & \texttimes & \checkmark & \texttimes
  & \checkmark & \texttimes & \checkmark & \texttimes & \texttimes & \checkmark \\
MetaDrive~\cite{metadrive}
  & \checkmark & $\sim$     & \texttimes & \checkmark & $\sim$
  & \checkmark & \texttimes & \texttimes & \texttimes & \texttimes & \checkmark \\
VISTA~\cite{vista}
  & $\sim$     & \texttimes & \texttimes & \checkmark & $\sim$
  & \checkmark & \texttimes & \texttimes & \texttimes & \texttimes & \checkmark \\
\midrule
\multicolumn{12}{@{}l}{\textit{Aerial / UAV}} \\
AirSim~\cite{airsim}
  & \texttimes & \texttimes & \checkmark & \checkmark & \checkmark
  & \checkmark & \texttimes & \checkmark & \texttimes & \checkmark & \checkmark \\
Flightmare~\cite{flightmare}
  & \texttimes & \texttimes & \checkmark & \checkmark & \checkmark
  & \checkmark & \texttimes & \texttimes & \texttimes & $\sim$ & \checkmark \\
FlightGoggles~\cite{flightgoggles}
  & \texttimes & \texttimes & \checkmark & \checkmark & \checkmark
  & \checkmark & \texttimes & \texttimes & \texttimes & $\sim$ & \checkmark \\
Gazebo / RotorS~\cite{gazebo,rotorsim}
  & $\sim$     & $\sim$     & \checkmark & \checkmark & $\sim$
  & \checkmark & \texttimes & \checkmark & \texttimes & \checkmark & \checkmark \\
OmniDrones~\cite{omnidrones}
  & \texttimes & \texttimes & \checkmark & \checkmark & \checkmark
  & \checkmark & \texttimes & \texttimes & \texttimes & \checkmark & \checkmark \\
gym-pybullet-drones~\cite{pybulletdrones}
  & \texttimes & \texttimes & \checkmark & \checkmark & $\sim$
  & \checkmark & \texttimes & \texttimes & \texttimes & $\sim$ & \checkmark \\
\midrule
\multicolumn{12}{@{}l}{\textit{Joint / Co-Simulation}} \\
TranSimHub~\cite{transimhub}
  & \checkmark & \checkmark & \checkmark & \texttimes & \texttimes
  & \texttimes & $\sim$     & ---        & \texttimes & \texttimes & \checkmark \\
CARLA+SUMO~\cite{carla,sumo}
  & \checkmark & \checkmark & \texttimes & \texttimes & \texttimes
  & $\sim$     & \texttimes & ---        & \texttimes & \texttimes & \checkmark \\
AirSim+Gazebo~\cite{airsim,gazebo,ros2}
  & $\sim$     & $\sim$     & \checkmark & \texttimes & \texttimes
  & $\sim$     & $\sim$     & ---        & \texttimes & $\sim$ & \checkmark \\
\midrule
\multicolumn{12}{@{}l}{\textit{Embodied AI \& RL}} \\
Isaac Lab~\cite{isaaclab}
  & \texttimes & \texttimes & $\sim$     & \checkmark & \checkmark
  & \checkmark & \texttimes & \texttimes & \checkmark & \checkmark & \checkmark \\
Isaac Gym~\cite{isaacgym}
  & \texttimes & \texttimes & $\sim$     & \checkmark & \checkmark
  & \checkmark & \texttimes & \texttimes & \texttimes & \texttimes & \checkmark \\
Habitat~\cite{habitat}
  & \texttimes & $\sim$     & \texttimes & \checkmark & \checkmark
  & \checkmark & \texttimes & \texttimes & \texttimes & \texttimes & \checkmark \\
SAPIEN~\cite{sapien}
  & \texttimes & \texttimes & \texttimes & \checkmark & \checkmark
  & \checkmark & \texttimes & \texttimes & \texttimes & $\sim$ & \checkmark \\
RoboSuite~\cite{robosuite}
  & \texttimes & \texttimes & \texttimes & \checkmark & \checkmark
  & \checkmark & \texttimes & \texttimes & \texttimes & $\sim$ & \checkmark \\
\midrule
\textbf{CARLA-Air (Ours)}
  & \checkmark & \checkmark\textsuperscript{$\dagger$} & \checkmark
  & \checkmark & \checkmark & \checkmark & \checkmark
  & \checkmark & \checkmark & \checkmark & \checkmark \\
\bottomrule
\end{tabular}

\vspace{3pt}
\raggedright
\scriptsize
\textsuperscript{$\dagger$}Pedestrian AI is inherited from CARLA and fully
functional; behavior under high actor density in joint scenarios is an active
engineering target (Section~\ref{sec:limitations}).
\end{table*}

\begin{table}[t]
\centering
\caption{%
  Architectural comparison of representative joint and co-simulation
  platforms.\textsuperscript{a}%
}
\label{tab:related_arch}
\footnotesize
\setlength{\tabcolsep}{3.2pt}
\renewcommand{\arraystretch}{1.05}
\begin{tabular}{@{}p{2.75cm}ccccc@{}}
\toprule
\textbf{Platform} &
\textbf{\shortstack{Single\\Proc.}} &
\textbf{\shortstack{API\\Kept}} &
\textbf{\shortstack{Shared\\Render}} &
\textbf{\shortstack{Sync\\Mode}} &
\textbf{\shortstack{Open\\Source}} \\
\midrule
TranSimHub~\cite{transimhub}
  & \texttimes & \texttimes & \texttimes & Msg. & \checkmark \\
CARLA+SUMO~\cite{carla,sumo}
  & \texttimes & $\sim$ & \texttimes & RPC & \checkmark \\
AirSim+Gazebo~\cite{airsim,gazebo,ros2}
  & \texttimes & $\sim$ & \texttimes & Decpl. & \checkmark \\
AirSim+CARLA\textsuperscript{b}
  & \texttimes & \texttimes & \texttimes & Decpl. & \checkmark \\
\midrule
\textbf{CARLA-Air (Ours)}
  & \checkmark & \checkmark & \checkmark & Shared & \checkmark \\
\bottomrule
\end{tabular}

\vspace{4pt}
\raggedright
\scriptsize
\textsuperscript{a}%
Single Proc.\,=\,single-process execution;
API Kept\,=\,preservation of native simulator APIs;
Shared Render\,=\,shared rendering pipeline;
Sync Mode: Msg.\,=\,message passing, Decpl.\,=\,decoupled execution,
Shared\,=\,shared-tick within one process.\\
\textsuperscript{b}%
Refers to community bridge solutions that run AirSim and CARLA as independent
processes connected via ROS\,2 or custom middleware; not an official product of
either project.
\end{table}

\section{System Architecture}
\label{sec:architecture}

\begin{figure}[!htbp]
\centering
\adjustbox{max width=\columnwidth, max height=0.32\textheight}{
\begin{tikzpicture}[
  font=\small,
  clientbox/.style={
    draw=gray!55, fill=gray!10, rounded corners=4pt,
    minimum width=3.2cm, minimum height=0.80cm,
    align=center, thick
  },
  rpcbox/.style={
    draw=#1, fill=white, rounded corners=4pt,
    minimum width=3.5cm, minimum height=0.72cm,
    align=center, thick
  },
  subbox/.style={
    draw=#1!50, fill=#1!7, rounded corners=5pt,
    minimum width=4.0cm, minimum height=1.80cm,
    align=center, thick
  },
  renbox/.style={
    draw=purple!50, fill=purple!7, rounded corners=4pt,
    minimum width=10.4cm, minimum height=0.72cm,
    align=center, thick
  },
  arr/.style={->, >=stealth, thick},
  darr/.style={<->, >=stealth, thick},
]

\node[clientbox] (cc) at (-2.6, 6.9) {Python\\CARLA Client};
\node[clientbox] (ac) at ( 2.6, 6.9) {Python\\AirSim Client};

\draw[draw=gray!40, dashed, rounded corners=8pt, line width=1.2pt]
  (-5.4, -1.7) rectangle (5.4, 6.2);
\node[font=\footnotesize\itshape, text=gray!55, anchor=north west]
  at (-5.4, 6.2) {Single UE4 Process};

\node[rpcbox=teal]   (rc) at (-2.6, 5.4) {CARLA RPC Server};
\node[rpcbox=orange] (ra) at ( 2.6, 5.4) {AirSim RPC Server};

\draw[draw=blue!40, fill=blue!2, rounded corners=6pt, thick]
  (-5.1, 0.6) rectangle (5.1, 4.6);
\node[font=\small\bfseries, text=blue!65, anchor=north] at (0, 4.6)
  {CARLA-Air Game Mode\enspace
   {\normalfont\scriptsize\color{blue!45}(\texttt{CARLAAirGameMode})}};

\node[subbox=blue] (gs) at (-2.55, 2.45) {
  \textbf{Ground Subsystems}\\[4pt]
  {\scriptsize episode\;\textbullet\;weather\;\textbullet\;traffic}\\
  {\scriptsize actors\;\textbullet\;scenario recorder}\\[6pt]
  {\tiny\itshape acquired via inheritance}
};

\node[subbox=orange] (af) at (2.55, 2.45) {
  \textbf{Aerial Flight Actor}\\[4pt]
  {\scriptsize physics engine\;\textbullet\;flight pawn}\\
  {\scriptsize aerial sensor suite}\\[6pt]
  {\tiny\itshape composed in \textsc{BeginPlay}}
};

\node[renbox] (ren) at (0, -1.1) {
  Shared UE4 Rendering Pipeline\enspace
  {\scriptsize\color{purple!60}RGB\;\textbullet\;depth\;\textbullet\;%
   segmentation\;\textbullet\;weather effects}
};

\draw[darr, gray!60] (cc.south) -- (rc.north);
\draw[darr, gray!60] (ac.south) -- (ra.north);
\draw[arr, teal!70]   (rc.south) -- (-2.6, 4.6);
\draw[arr, orange!70] (ra.south) -- ( 2.6, 4.6);
\draw[arr, gray!55] (0, 0.6) -- (ren.north);

\end{tikzpicture}
}
\caption{%
  Runtime architecture of CARLA-Air.
  A single engine process hosts both simulation backends, each communicating
  with its respective Python client via an independent RPC server.
  \texttt{CARLAAirGameMode} acquires ground simulation functionality through
  class inheritance and integrates the aerial flight actor through composition.
  All world actors share a single rendering pipeline.%
}
\label{fig:arch_overview}
\end{figure}
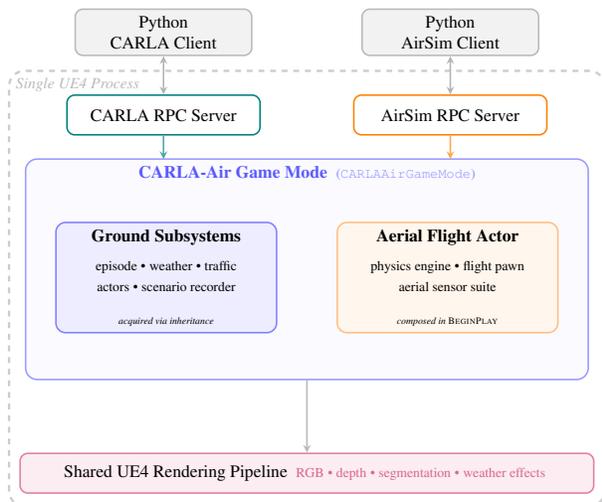

CARLA-Air integrates CARLA~\cite{carla} and AirSim~\cite{airsim} within a
single Unreal Engine~\cite{ue4} process through a minimal bridging layer that
resolves a fundamental engine-level initialization conflict while preserving
both platforms' native APIs, physics engines, and rendering pipelines intact.
Fig.~\ref{fig:arch_overview} presents the high-level runtime structure; the
following subsections elaborate each design decision.


\subsection{Plugin and Dependency Structure}
\label{sec:arch:plugins}

The system comprises two plugin modules that load sequentially during engine
startup.
The ground simulation plugin initializes first, establishing its world
management subsystems before any game logic executes.
The aerial simulation plugin declares a compile-time dependency on the ground
plugin, enabling \texttt{CARLAAirGameMode} to access the ground platform's
initialization interfaces during its own startup phase.
This dependency is strictly one-directional: no ground-platform source file
references any aerial component, preserving the upstream CARLA codebase's
update path without modification.
Two independent RPC servers run concurrently within the single
process---one per simulator---allowing the native Python clients of each
platform to connect without modification.

Version compatibility across the two upstream codebases, network configuration,
and port assignments are documented in Appendix~\ref{app:config}.

\subsection{The GameMode Conflict and Its Resolution}
\label{sec:arch:gamemode}

UE4 enforces a strict invariant: each world may have exactly one active game
mode.
CARLA's game mode orchestrates episode management, weather control, traffic
simulation, the actor lifecycle, and the RPC interface through a deep
inheritance chain.
AirSim's game mode performs a separate startup sequence---reading configuration
files, adjusting rendering settings, and spawning its flight actor.
Because the two game modes are unrelated by inheritance, assigning either to the
world map silently skips the other's initialization, rendering a large portion
of its API surface inoperative.

\paragraph{Architectural asymmetry.}
A structural difference between the two systems makes resolution tractable and
constitutes the central design insight of CARLA-Air.
CARLA's subsystems are tightly coupled to its game mode through inheritance and
privileged class relationships; they cannot be relocated outside the game mode
slot without invasive upstream refactoring.
AirSim's flight logic, by contrast, resides in a class derived from the
generic \emph{actor} base---not the game mode base---and can therefore be
spawned as a regular world actor at any point after world initialization.

\begin{figure*}[!htbp]
\centering
\adjustbox{max width=\textwidth, max height=0.3\textheight}{
\begin{tikzpicture}[
  font=\small,
  cbox/.style={
    draw=#1, fill=white, rounded corners=3pt,
    minimum width=3.1cm, minimum height=0.65cm,
    align=center, thick
  },
  slotbox/.style={
    draw=gray!70, fill=gray!8, rounded corners=4pt,
    minimum width=3.6cm, minimum height=0.72cm,
    align=center, thick
  },
  subbox/.style={
    draw=#1!50, fill=#1!7, rounded corners=3pt,
    minimum width=3.4cm, minimum height=0.72cm,
    align=center, font=\scriptsize, thick
  },
  inh/.style={->, >=open triangle 60, thick},
  comp/.style={->, >=stealth, dashed, thick, orange!75!black},
  cross/.style={red!65!black, very thick, font=\large\bfseries},
]

\node[font=\footnotesize\bfseries, text=red!65] at (-3.4, 4.3)
  {(a)~Naive approach};

\node[slotbox] (slot) at (-3.4, 3.55) {UE4 Game Mode Slot};

\node[cbox=blue]   (cgm) at (-5.0, 2.45) {CARLA\\Game Mode};
\node[cbox=orange] (agm) at (-1.8, 2.45) {AirSim\\Game Mode};

\draw[inh, blue!60]   (cgm.north) -- (-5.0, 3.20);
\draw[inh, orange!70] (agm.north) -- (-1.8, 3.20);
\node[cross] at (-5.0, 3.20) {\texttimes};
\node[cross] at (-1.8, 3.20) {\texttimes};

\node[subbox=blue]   at (-5.0, 1.42) {Ground subsystems};
\node[subbox=orange] at (-1.8, 1.42) {Aerial flight actor};
\draw[->, red!55, dashed, thick] (cgm.south) -- ++(0,-0.40);
\draw[->, red!55, dashed, thick] (agm.south) -- ++(0,-0.40);

\node[font=\scriptsize, text=red!65, align=center] at (-3.4, 0.60)
  {One mode silently discarded;\\corresponding API inoperative};

\node[font=\footnotesize\bfseries, text=green!45!black] at (3.6, 4.3)
  {(b)~CARLA-Air solution};

\node[cbox=gray!70, minimum width=3.6cm] (base) at (3.6, 3.55)
  {CARLA Game Mode Base};

\node[cbox=blue, minimum width=3.8cm] (unified) at (3.6, 2.45)
  {\texttt{CARLAAirGameMode}};
\node[font=\tiny\itshape, text=blue!55, anchor=north] at (3.6, 2.12)
  {occupies game mode slot};

\draw[inh, gray!60] (unified.north) -- (base.south);

\node[subbox=blue,   minimum width=2.8cm] (gss) at (2.1, 1.15)
  {Ground Subsystems\\[-1pt]{\tiny\itshape inherited}};
\node[subbox=orange, minimum width=2.8cm] (afs) at (5.1, 1.15)
  {Aerial Flight Actor\\[-1pt]{\tiny\itshape composed}};

\draw[->, blue!60, thick] (unified.south) -| (gss.north);
\draw[comp]               (unified.south) -| (afs.north);

\end{tikzpicture}
}
\caption{%
  Resolving the UE4 single-game-mode constraint.
  (a)~Both backends provide independent game mode classes; assigning either
  silently discards the other.
  (b)~\texttt{CARLAAirGameMode} inherits all ground functionality from CARLA's
  game mode base while composing the aerial flight actor as a spawned world
  actor.%
}
\label{fig:gamemode_design}
\end{figure*}
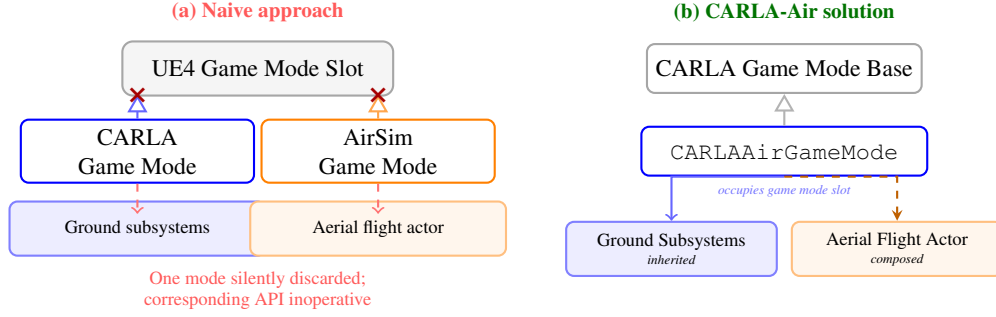

\paragraph{Solution: single inheritance plus composition.}
We introduce \texttt{CARLAAirGameMode}, which inherits from CARLA's game mode
base and occupies the single available slot.
All ground simulation subsystems are thereby acquired through the standard UE4
lifecycle.
The aerial flight actor is then \emph{composed} into the world during the
engine's \textsc{BeginPlay} phase, after ground initialization is complete, and
never competes for the game mode slot.
Fig.~\ref{fig:gamemode_design} contrasts the naive conflict with this adopted
solution.


\paragraph{Source footprint.}
The integration modifies exactly two files in the upstream CARLA source tree:
two previously private members are promoted to protected visibility and one
privileged class declaration is added.
All remaining integration code resides within the aerial plugin as purely
additive content.
The complete modification summary is provided in Appendix~\ref{app:source_mods}.

\subsection{Coordinate System Mapping}
\label{sec:arch:coords}

\begin{figure}[!htbp]
\centering
\includegraphics[width=\columnwidth]{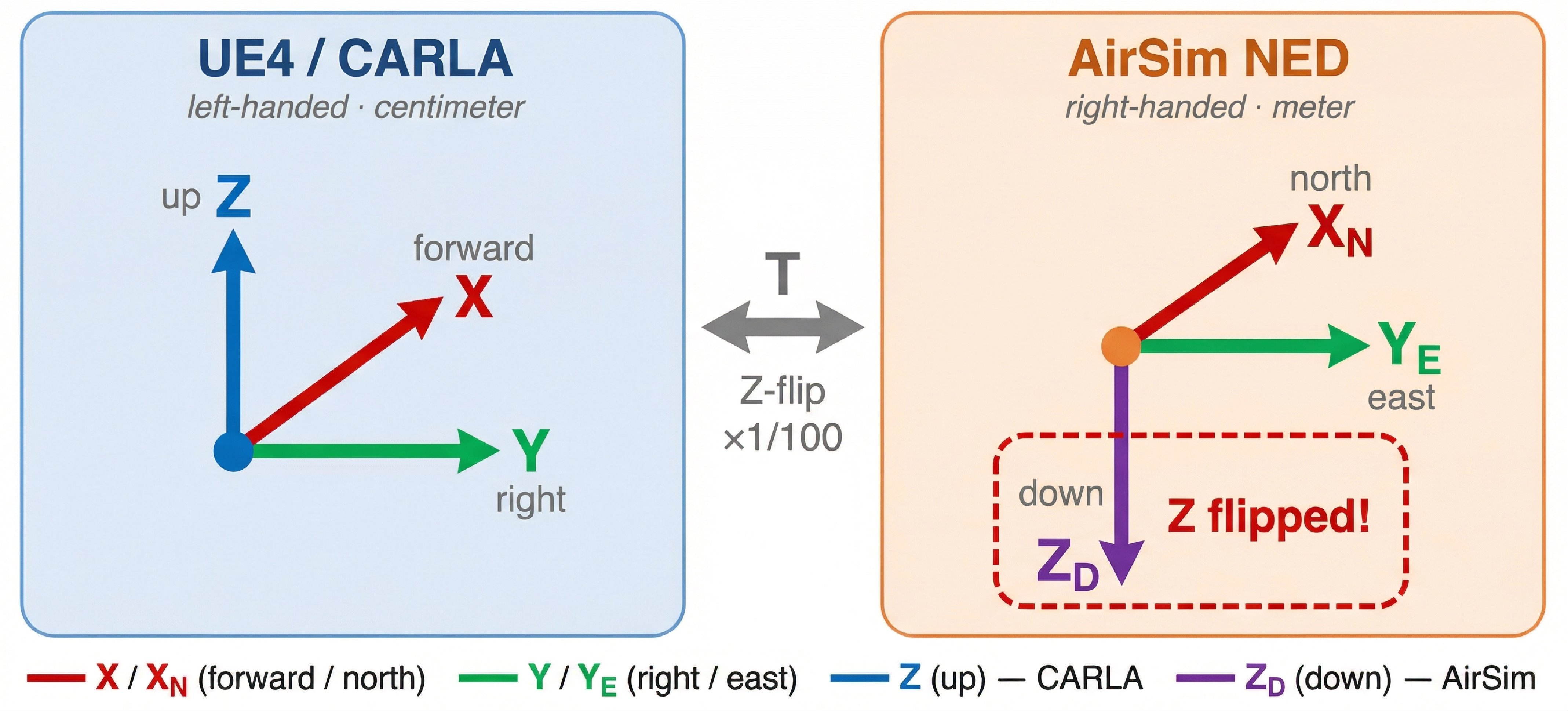}
\caption{%
  Coordinate frames of the two simulation backends.
  The transform~$T$ requires only a Z-axis sign flip and a
  centimeter-to-meter scale factor; the forward ($X$/$X_N$) and
  rightward ($Y$/$Y_E$) axes are aligned across both conventions.%
}
\label{fig:coord_frames}
\end{figure}

CARLA and AirSim employ incompatible spatial reference frames that must be
reconciled to co-register aerial and ground sensor data.
CARLA inherits UE4's left-handed system with X forward, Y right, and Z up, in
centimeters.
AirSim adopts a right-handed North-East-Down (NED) frame with X north, Y east,
and Z down, in meters.
Fig.~\ref{fig:coord_frames} illustrates both frames and their geometric
relationship.


Let $\mathbf{p} \in \mathbb{R}^3$ denote a point in the UE4 world frame and
$\mathbf{o}$ the shared world origin established during initialization.
The equivalent NED position is
\begin{equation}
  \mathbf{p}_{\mathrm{NED}}
  \;=\; \frac{1}{100}
  \begin{pmatrix} p_x - o_x \\ p_y - o_y \\ -(p_z - o_z) \end{pmatrix},
  \label{eq:pos_transform}
\end{equation}
\noindent where the scale factor converts centimeters to meters and the sign
reversal on the third component reflects the Z-axis inversion.
Because the X and Y axes are directionally aligned, no axis permutation is
required.

For orientation, let $q = (w, q_x, q_y, q_z)$ denote a unit quaternion in the
UE4 frame.
The equivalent NED quaternion is
\begin{equation}
  q_{\mathrm{NED}} \;=\; \bigl(w,\; q_x,\; q_y,\; -q_z\bigr),
  \label{eq:rot_transform}
\end{equation}
\noindent where negating $q_z$ accounts for the Z-axis reversal and the
associated change of frame handedness.
Eqs.~(\ref{eq:pos_transform}) and~(\ref{eq:rot_transform}) together fully
specify the pose transform, enabling consistent fusion of drone attitude from
the aerial API with vehicle heading from the ground API across all joint
simulation workflows.

\subsection{Asset Import Pipeline}
\label{sec:arch:assets}

CARLA-Air provides an extensible asset import pipeline that allows researchers
to bring custom robot platforms, UAV models, vehicles, and environment assets
into the shared simulation world. Imported assets are fully integrated into the
joint simulation environment: they participate in the same physics tick and
rendering pass as all built-in actors, respond to both ground and aerial API
calls, and are visible to all sensor modalities across both simulation
backends. This capability enables evaluation of custom hardware designs---such
as novel multirotor configurations or application-specific ground
robots---within realistic air-ground scenarios without modifying the core
CARLA-Air codebase. Fig.~\ref{fig:custom_assets} shows two examples of
user-imported assets operating within the platform.


\section{Performance Evaluation}
\label{sec:perf}

This section evaluates CARLA-Air under representative joint air-ground
workloads across three experiments: frame-rate and resource scaling
(Section~\ref{sec:perf:fps}), memory stability under sustained operation
(Section~\ref{sec:perf:vram}), and communication latency
(Section~\ref{sec:perf:rpc}).
Full configuration parameters and raw data are deferred to
Appendix~\ref{app:config}.

\paragraph{Reference hardware.}
All measurements are collected on a workstation equipped with an NVIDIA
RTX~A4000 (16\,GB GDDR6), AMD Ryzen~7 5800X (8-core, 4.7\,GHz), and 32\,GB
DDR4-3200, running Ubuntu~20.04~LTS.
The simulator runs in Epic quality mode with Town10HD loaded unless stated
otherwise.
All aerial experiments use the built-in SimpleFlight controller with default
PID gains.
CPU affinity and GPU power limits are left at system defaults to reflect
realistic research deployment conditions.

\subsection{Benchmark Methodology}
\label{sec:perf:method}

Reliable performance measurement requires eliminating startup
transients---map-loading jitter, first-frame shader compilation, and actor
lifecycle initialization must be discarded before steady-state sampling begins.
Algorithm~\ref{alg:bench} formalizes the benchmark harness used throughout this
section.


\begin{algorithm}[!htbp]
\footnotesize
\caption{Simulation Performance Benchmark Harness}
\label{alg:bench}
\begin{algorithmic}[1]
\Require Workload $\mathcal{W}$, warm-up ticks $T_w$, measurement ticks $T_m$
\Ensure Frame-time $\mathbf{f}$, VRAM $\mathbf{v}$, latency $\boldsymbol{\ell}$
\State Connect to simulation; load map; wait for shader compilation
\State Spawn actors/sensors per $\mathcal{W}$; enable synchronous mode
\For{$t = 1$ \textbf{to} $T_w$} \Comment{Warm-up---discarded}
    \State Advance one tick
\EndFor
\State $\mathbf{f}, \mathbf{v}, \boldsymbol{\ell} \leftarrow \langle\rangle$
\For{$t = 1$ \textbf{to} $T_m$}
    \State $t_0 \leftarrow \textsc{Now}()$; advance one tick
    \State $\mathbf{f}.\textsc{Append}(1/(\textsc{Now}() - t_0))$
    \If{VRAM sample interval reached}
        \State Record GPU memory $\to \mathbf{v}$; measure API round-trip $\to \boldsymbol{\ell}$
    \EndIf
\EndFor
\State Destroy all actors; disable synchronous mode
\State \Return $\mathbf{f}$, $\mathbf{v}$, $\boldsymbol{\ell}$
\end{algorithmic}
\end{algorithm}

Each profile uses $T_w = 200$ warm-up ticks and $T_m = 2{,}000$ measurement
ticks.
VRAM is sampled every 60\,s.
All reported frame rates are the harmonic mean of $\mathbf{f}$---the
appropriate central tendency for rate quantities---with standard deviations
alongside.
The latency benchmark issues 500 warm-up calls followed by 5\,000 measurement
calls; actor spawn calls are each paired with an immediate destroy to prevent
scene-state accumulation from contaminating subsequent measurements.

\subsection{Experiment~1: Frame Rate and Resource Scaling}
\label{sec:perf:fps}

\paragraph{Workload design.}
Under synchronous-mode operation, per-tick wall time is bounded by the slowest
of three concurrent contributors: the rendering thread (GPU-bound), the
ground-actor dispatch loop (CPU-bound), and the aerial physics engine
(CPU-bound, asynchronous at ${\approx}1{,}000$\,Hz on a dedicated thread).
Sensor rendering dominates at high resolution due to GPU memory bandwidth
saturation; actor population contributes via per-mesh draw calls.
These relationships motivate the workload stratification in
Table~\ref{tab:fps_scaling}.


\begin{table*}[!htbp]
\centering
\caption{%
  Frame rate and resource consumption across representative joint workloads
  (RTX~A4000, Town10HD, Epic quality, synchronous mode).
  Harmonic mean $\pm$ 1\,SD over 2\,000 ticks after 200 warm-up.%
}
\label{tab:fps_scaling}
\footnotesize
\setlength{\tabcolsep}{6pt}
\renewcommand{\arraystretch}{1.10}
\begin{tabular}{@{}llccc@{}}
\toprule
\textbf{Profile} & \textbf{Configuration} & \textbf{FPS} & \textbf{VRAM (MiB)} & \textbf{CPU (\%)} \\
\midrule
\multicolumn{5}{@{}l}{\textit{Standalone baselines}} \\
Ground sim only\textsuperscript{s}
  & 3 vehicles + 2 pedestrians; 8 sensors @ $1280\!\times\!720$
  & $28.4 \pm 1.2$ & $3{,}821 \pm 10$ & $31 \pm 3$ \\
Aerial sim only\textsuperscript{s}
  & 1 drone; 8 sensors @ $1280\!\times\!720$
  & $44.7 \pm 2.1$ & $2{,}941 \pm 8\phantom{0}$ & $29 \pm 3$ \\
\midrule
\multicolumn{5}{@{}l}{\textit{Joint workloads}} \\
Idle
  & Town10HD; no actors; no sensors
  & $60.0 \pm 0.4$ & $3{,}702 \pm 8\phantom{0}$ & $12 \pm 2$ \\
Ground only
  & 3 vehicles + 2 pedestrians; 8 sensors @ $1280\!\times\!720$
  & $26.3 \pm 1.4$ & $3{,}831 \pm 11$ & $38 \pm 4$ \\
Moderate joint
  & 3 vehicles + 2 pedestrians + 1 drone; 8 sensors @ $1280\!\times\!720$
  & $19.8 \pm 1.1$ & $3{,}870 \pm 13$ & $54 \pm 5$ \\
Traffic surveillance
  & 8 autopilot vehicles + 1 drone; 1 aerial RGB @ $1920\!\times\!1080$
  & $20.1 \pm 1.8$ & $3{,}874 \pm 15$ & $61 \pm 6$ \\
Stability endurance
  & Moderate joint; 357 spawn/destroy cycles; 3\,hr continuous
  & $19.7 \pm 1.3$ & $3{,}878 \pm 17$ & $55 \pm 5$ \\
\bottomrule
\end{tabular}

\vspace{3pt}
\raggedright\scriptsize
\textsuperscript{s}Standalone baseline: single simulator running without CARLA-Air integration.
\end{table*}

\paragraph{Analysis.}
The moderate joint configuration---combining ground traffic, an aerial agent,
and a full sensor suite---sustains $19.8 \pm 1.1$\,FPS, sufficient for
closed-loop policy evaluation at standard RL episode lengths.
Comparing the standalone ground baseline ($28.4$\,FPS) against the moderate
joint profile ($19.8$\,FPS) quantifies total integration overhead at
$8.6$\,FPS (30.3\%), of which $2.1$\,FPS is attributable to ground co-hosting
and the remaining $6.5$\,FPS to the aerial physics engine.
Crucially, this aerial overhead manifests entirely in CPU utilization ($54\%$
vs.\ $38\%$) rather than GPU memory: VRAM differs by only 39\,MiB between
ground-only and moderate-joint profiles.
The traffic surveillance profile retains comparable throughput ($20.1$\,FPS)
despite doubling the vehicle count, confirming that sensor rendering---not
actor population---is the dominant cost driver at $1920\times1080$.
The 3-hour endurance profile ($19.7 \pm 1.3$\,FPS) confirms that throughput
does not degrade under sustained operation.

\subsection{Experiment~2: Memory Stability}
\label{sec:perf:vram}

\paragraph{Steady-state VRAM profile.}
The base map asset load accounts for approximately 3\,702\,MiB at idle
(Town10HD, Epic quality); each additional ground sensor contributes
4--10\,MiB at $1280\times720$.
A transient peak of approximately 5\,000\,MiB occurs during map loading and
resolves within 30\,s.
All steady-state profiles remain within 3\,878\,MiB, retaining approximately
12\,506\,MiB (76\%) of the 16\,GB device budget for co-located workloads such
as GPU-based policy training.

\begin{figure}[!htbp]
\centering
\adjustbox{max width=\columnwidth, max height=0.22\textheight}{
\begin{tikzpicture}[font=\footnotesize]


\draw[->, thick, black!70] (0,0) -- (7.6,0) node[right] {\scriptsize Time (min)};
\draw[->, thick, black!70] (0,0) -- (0,4.5) node[above] {};
\node[rotate=90, font=\scriptsize, anchor=south] at (-0.85, 2.0) {VRAM (MiB)};

\foreach \yc/\ylbl in {0/3800, 1.333/3840, 2.667/3880, 4.0/3920}{
  \draw (-0.08, \yc) -- (0, \yc);
  \draw[black!15, dashed] (0, \yc) -- (7.3, \yc);
  \node[left, font=\tiny] at (-0.1, \yc) {\ylbl};
}

\foreach \xc/\xlbl in {0/0, 1.167/30, 2.333/60, 3.5/90, 4.667/120, 5.833/150, 7.0/180}{
  \draw (\xc, -0.08) -- (\xc, 0);
  \node[below, font=\tiny] at (\xc, -0.1) {\xlbl};
}

\draw[blue!65!black, semithick, line join=round]
  plot[smooth] coordinates {
    (0.0, 2.267)
    (0.389, 2.667)
    (0.778, 1.900)
    (1.167, 2.400)
    (1.556, 2.033)
    (1.944, 2.500)
    (2.333, 2.133)
    (2.722, 2.600)
    (3.111, 2.300)
    (3.500, 2.533)
    (3.889, 2.367)
    (4.278, 2.733)
    (4.667, 2.433)
    (5.056, 2.667)
    (5.444, 2.500)
    (5.833, 2.767)
    (6.222, 2.533)
    (6.611, 2.733)
    (7.0,   2.600)
  };

\draw[black!30, dotted, semithick] (1.167, 0) -- (1.167, 4.2);
\draw[black!30, dotted, semithick] (5.833, 0) -- (5.833, 4.2);
\node[font=\tiny, text=black!50] at (0.58, 4.1) {early};
\node[font=\tiny, text=black!50] at (6.42, 4.1) {late};

\draw[orange!75!black, thin, dashed] (0.0, 2.267) -- (1.167, 2.267);
\node[right, font=\tiny, text=orange!65!black] at (1.2, 2.267)
  {$\bar{v}_{\text{early}}\!=\!3868$};

\draw[orange!75!black, thin, dashed] (5.833, 2.600) -- (7.3, 2.600);
\node[right, font=\tiny, text=orange!65!black] at (7.32, 2.600)
  {$\bar{v}_{\text{late}}\!=\!3878$};

\draw[blue!65!black, semithick] (0.2, 4.3) -- (0.65, 4.3);
\node[right, font=\tiny] at (0.7, 4.3) {VRAM (60\,s interval)};
\draw[orange!75!black, thin, dashed] (3.5, 4.3) -- (3.95, 4.3);
\node[right, font=\tiny] at (4.0, 4.3) {Phase mean};

\end{tikzpicture}
}
\caption{%
  VRAM trace over a 3-hour stability run (357 spawn/destroy cycles, moderate
  joint configuration, RTX~A4000).
  Early-to-late drift is ${\approx}10$\,MiB; linear regression yields
  $R^2 = 0.11$, confirming no significant memory accumulation.%
}
\label{fig:vram_timeseries}
\end{figure}
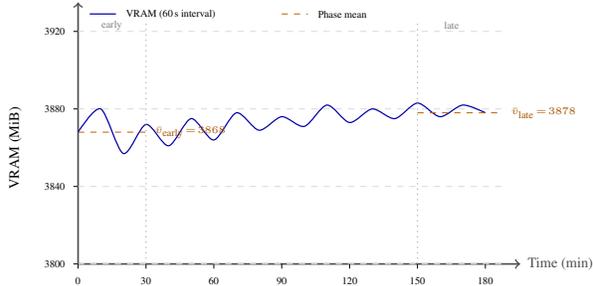

\paragraph{Leak verification.}
Table~\ref{tab:stability} summarises the 3-hour endurance run across 357 actor
spawn-and-destroy cycles.
A linear regression of VRAM against cycle index yields a slope of
$0.49$\,MiB/cycle with $R^2 = 0.11$, indicating no statistically significant
accumulation trend.
Fig.~\ref{fig:vram_timeseries} visualises the VRAM trace: the negligible
early-to-late drift ($3{,}868 \to 3{,}878$\,MiB, approximately 10\,MiB over
327 cycles) is attributable to residual render-target caching rather than
lifecycle leakage.


\begin{table}[!htbp]
\centering
\caption{%
  Stability endurance results over 3\,hours and 357 actor lifecycle cycles
  (moderate joint configuration).%
}
\label{tab:stability}
\small
\setlength{\tabcolsep}{3pt}
\renewcommand{\arraystretch}{1.08}
\resizebox{\columnwidth}{!}{%
\begin{tabular}{@{}lcc@{}}
\toprule
\textbf{Metric} & \textbf{Early (cycles 1--30)} & \textbf{Late (cycles 328--357)} \\
\midrule
Frame rate (FPS)       & $19.9 \pm 1.2$   & $19.7 \pm 1.3$   \\
VRAM (MiB)             & $3{,}868 \pm 14$ & $3{,}878 \pm 17$ \\
CPU utilization (\%)   & $53 \pm 5$       & $55 \pm 5$       \\
API error count        & 0                & 0                \\
Crash count            & 0                & 0                \\
VRAM regression slope  & \multicolumn{2}{c}{$0.49$\,MiB/cycle, $R^2 = 0.11$} \\
\bottomrule
\end{tabular}}%
\end{table}

Zero API errors and zero simulation crashes across 357 cycles validate the
robustness of the joint environment under the repeated agent-reset patterns
typical of reinforcement-learning training.

\subsection{Experiment~3: Communication Latency}
\label{sec:perf:rpc}

Both simulation APIs operate within the same process on the loopback interface,
eliminating inter-process serialization overhead.
The two RPC servers use distinct wire protocols and port assignments (detailed
in Appendix~\ref{app:config}).
Table~\ref{tab:rpc_latency} reports round-trip latency for representative API
calls.

\begin{table*}[!t]
\centering
\caption{%
  Round-trip API call latency on the loopback interface
  (median $\pm$ IQR; 5\,000 calls after 500 warm-up; RTX~A4000; idle scene).%
}
\label{tab:rpc_latency}
\small
\setlength{\tabcolsep}{8pt}
\renewcommand{\arraystretch}{1.0}
\begin{tabular}{@{}llrr@{}}
\toprule
\textbf{API} & \textbf{Call} & \textbf{Median ($\mu$s)} & \textbf{IQR ($\mu$s)} \\
\midrule
Ground sim  & World state snapshot            & 320        & 40  \\
Ground sim  & Actor transform query           & 280        & 35  \\
Ground sim  & Actor spawn (+ paired destroy)  & 1{,}850    & 210 \\
Ground sim  & Actor destroy                   & 920        & 95  \\
Aerial sim  & Multirotor state query          & 410        & 55  \\
Aerial sim  & Image capture (1 RGB stream)    & 3{,}200    & 380 \\
Aerial sim  & Velocity command dispatch       & 490        & 60  \\
\midrule
Bridge IPC~\cite{transimhub}
            & Cross-process state sync (ref.) & 3{,}000    & 2{,}000 \\
\bottomrule
\end{tabular}
\vspace{-2mm}
\end{table*}

\paragraph{Analysis.}
Lightweight state queries complete in 280--490\,$\mu$s, well below the per-tick
budget at 20\,FPS (50\,ms), confirming that API overhead does not contribute
meaningfully to frame-time variance.
Actor spawn latency (1\,850\,$\mu$s) reflects a one-time GPU synchronization
point for render-asset registration at episode reset.
Image capture latency (3\,200\,$\mu$s) covers the full sensor
pipeline---rendering, buffer readback, and serialization---and is overlapped
with the rendering thread at synchronous tick rates.
All measured values fall below the lower bound of bridge-based cross-process
synchronization costs reported in~\cite{transimhub}
(1\,000--5\,000\,$\mu$s per frame).

\paragraph{Tick-rate reconciliation.}
The aerial physics engine advances at ${\approx}1{,}000$\,Hz on its dedicated
thread, while the rendering tick runs at ${\approx}20$\,Hz under the moderate
joint workload.
Sensor callbacks read the aerial state at rendering tick boundaries, so each
ground-aerial sensor frame pair reflects the drone's integrated physical state
over ${\approx}50$ aerial physics steps.
Ground actor states are also resolved at tick boundaries, ensuring temporal
co-registration across all sensor modalities within a single tick.
Applications requiring finer aerial state resolution should reduce the fixed
simulation delta-time accordingly; the throughput trade-off follows from
Table~\ref{tab:fps_scaling}.

\section{Representative Applications}
\label{sec:apps}

CARLA-Air is validated through five representative workflows that collectively
exercise the platform's core capabilities across the four research directions
identified in Section~\ref{sec:introduction}: air-ground cooperation, embodied
navigation and vision-language action, multi-modal perception and dataset
construction, and reinforcement-learning-based policy training.
Table~\ref{tab:workflows_summary} provides a structured overview; detailed
descriptions follow in Sections~\ref{sec:apps:coop}--\ref{sec:apps:rl}.


\begin{table*}[!htbp]
\centering
\caption{%
  Summary of five representative workflows validated on CARLA-Air,
  mapped to the four research directions from Section~\ref{sec:introduction}.%
}
\label{tab:workflows_summary}
\footnotesize
\setlength{\tabcolsep}{4pt}
\renewcommand{\arraystretch}{1.10}
\begin{tabular}{@{}clp{2.8cm}p{2.8cm}cp{3.2cm}@{}}
\toprule
& \textbf{Workflow} & \textbf{Research Direction} &
\textbf{Platform Features Exercised} & \textbf{FPS} &
\textbf{Key Outcome} \\
\midrule
W1 & Precision landing
   & Air-ground cooperation
   & Tick-sync control, descent planning, cross-frame coordination
   & ${\approx}19$
   & $<0.5$\,m final landing error \\
W2 & VLN/VLA data generation
   & Embodied navigation
   & Dual-view sensing, waypoint planning, semantic annotation
   & ---
   & Cross-view VLN data pipeline \\
W3 & Multi-modal dataset
   & Perception \& dataset
   & 12-stream sync capture, shared tick
   & ${\approx}17$
   & ${\leq}1$-tick alignment error \\
W4 & Cross-view perception
   & Perception \& dataset
   & Shared renderer, weather consistency
   & ${\approx}18$
   & 14/14 weather presets verified \\
W5 & RL training env.
   & RL policy training
   & Sync stepping, stable resets, cross-domain reward
   & ---
   & 357 reset cycles, 0 crashes \\
\bottomrule
\end{tabular}
\end{table*}

All workflows share a common dual-client architecture: both API clients execute
within the same Python process and operate on the same world state without
inter-process communication.
The following minimal pattern is common to every workflow:

{\footnotesize
\begin{verbatim}
ground = carla.Client('localhost', 2000)
aerial = airsim.MultirotorClient()
world  = ground.get_world()  # shared
aerial.enableApiControl(True)
\end{verbatim}
}


\begin{figure}[!htbp]
\centering
\adjustbox{max width=\columnwidth, max height=0.22\textheight}{
\begin{tikzpicture}[
  font=\footnotesize,
  box/.style={draw, rounded corners=3pt, minimum width=2.2cm,
              minimum height=0.56cm, align=center, fill=gray!10},
  boldbox/.style={draw, rounded corners=3pt, minimum width=4.8cm,
                  minimum height=0.56cm, align=center, fill=blue!12, thick},
  serverbox/.style={draw, rounded corners=3pt, minimum width=2.2cm,
                    minimum height=0.56cm, align=center, fill=orange!15},
  ue4box/.style={draw, rounded corners=3pt, minimum width=4.6cm,
                 minimum height=0.56cm, align=center, fill=green!10, thick},
  sharedbox/.style={draw, rounded corners=3pt, minimum width=4.6cm,
                    minimum height=0.56cm, align=center, fill=yellow!12},
  arrow/.style={-stealth, semithick},
  darrow/.style={stealth-stealth, semithick, dashed},
]

\node[boldbox] (pyproc) at (0, 3.6) {User Python Script};

\node[box] (gndC) at (-1.8, 2.6) {\scriptsize Ground Client};
\node[box] (airC) at ( 1.8, 2.6) {\scriptsize Aerial Client};

\node[serverbox] (gndR) at (-1.8, 1.7) {\scriptsize Ground RPC};
\node[serverbox] (airR) at ( 1.8, 1.7) {\scriptsize Aerial RPC};

\node[ue4box] (ue4) at (0, 0.85) {\scriptsize Unified UE4 Process};

\node[box, minimum width=1.8cm, fill=blue!8] (gndGM) at (-1.8, -0.05)
  {\scriptsize Ground\\[-1pt]Game Mode};
\node[box, minimum width=1.8cm, fill=blue!8] (airSM) at (1.8, -0.05)
  {\scriptsize Aerial\\[-1pt]Sim Mode};

\node[sharedbox] (shared) at (0, -0.95)
  {\scriptsize Shared World --- Actors, Physics, Rendering, Weather};

\draw[arrow] (pyproc.south) -- ++(0,-0.15) -| (gndC.north);
\draw[arrow] (pyproc.south) -- ++(0,-0.15) -| (airC.north);
\draw[darrow] (gndC.south) -- (gndR.north);
\draw[darrow] (airC.south) -- (airR.north);
\draw[arrow] (gndR.south) -- (gndGM.north);
\draw[arrow] (airR.south) -- (airSM.north);
\draw[arrow] (gndGM.south) -- (shared.north -| gndGM.south);
\draw[arrow] (airSM.south) -- (shared.north -| airSM.south);

\begin{pgfonlayer}{background}
  \node[draw=green!50!black, semithick, rounded corners=6pt, dashed,
        fit=(ue4)(gndGM)(airSM)(shared),
        inner sep=7pt, fill=green!3] {};
\end{pgfonlayer}

\end{tikzpicture}
}
\caption{%
  Dual-client architecture shared by all five workflows.
  A single Python script drives both clients, connected via TCP to independent
  RPC servers inside the unified engine process.%
}
\label{fig:workflow_arch}
\end{figure}
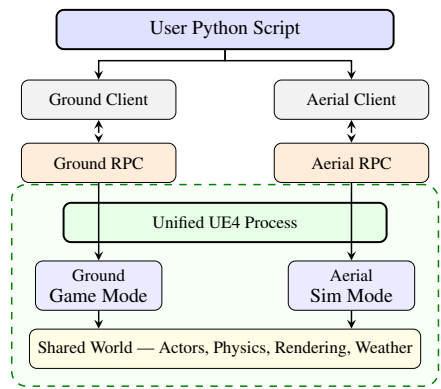

\subsection{W1: Air-Ground Cooperative Precision Landing}
\label{sec:apps:coop}

Autonomous precision landing of a UAV on a moving ground vehicle is a
representative and challenging scenario for air-ground cooperation, with
direct applications in drone-assisted logistics, mobile recharging, and
multi-agent coordination.
This workflow demonstrates real-time cross-domain coordination within
CARLA-Air's tick-synchronous control loop: the drone must continuously track
the vehicle's position, plan a descent trajectory, and execute a smooth
touchdown---all while the vehicle is in motion through urban traffic.

\paragraph{Setup.}
A ground vehicle follows an autopilot route through Town10HD at moderate
speed.
A drone starts at ${\approx}12$\,m altitude above the vehicle and is tasked
with landing on the vehicle's roof.
At each synchronous tick, the vehicle's 3D pose is queried via the ground API,
transformed into the aerial NED frame using the coordinate mapping from
Section~\ref{sec:arch:coords}, and used to compute a descent command issued
to the aerial flight controller.

\paragraph{Control architecture.}
The landing controller operates in three phases: \emph{approach}
(horizontal alignment with the vehicle), \emph{descent} (controlled altitude
reduction while maintaining horizontal tracking), and \emph{touchdown}
(final landing).
Let $\mathbf{q}_k \in \mathbb{R}^2$ denote the vehicle's horizontal position
at tick $k$ in the NED frame.
The drone's commanded horizontal target tracks the vehicle continuously:
\begin{equation}
\hat{\mathbf{q}}_k = \mathbf{q}_k + \mathbf{d},
\qquad
\hat{z}_k = z_k^{\mathrm{ref}},
\label{eq:landing_control}
\end{equation}
where $\mathbf{d}$ is the coordinate origin offset from
Eq.~\eqref{eq:coord_offset} and $z_k^{\mathrm{ref}}$ decreases according to
a smooth descent profile from the initial altitude to ground level.

\begin{figure*}[!htbp]
\centering
\includegraphics[width=\textwidth, height=0.38\textheight, keepaspectratio]{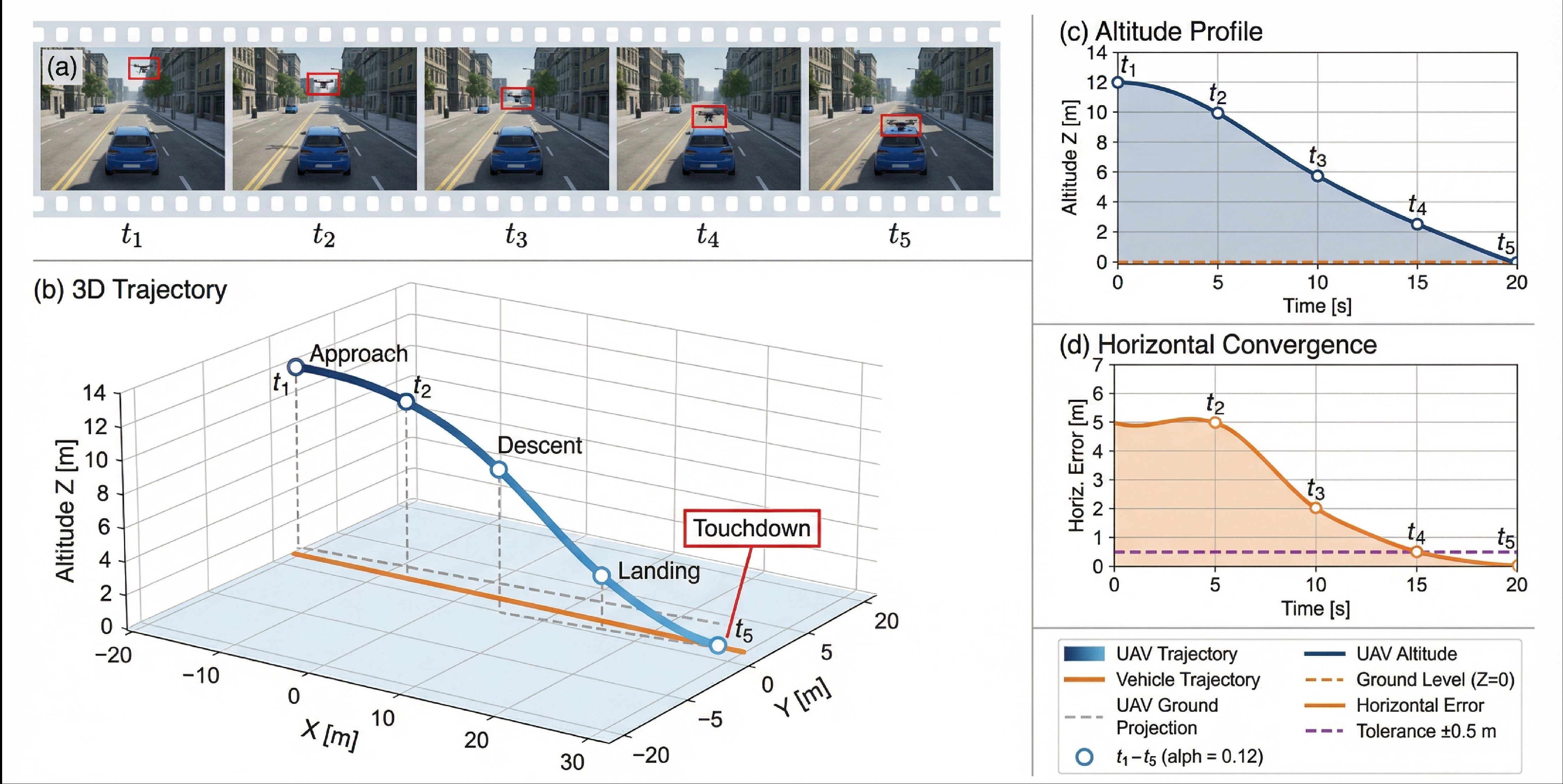}
\caption{%
  W1: Air-ground cooperative precision landing on a moving vehicle.
  \textbf{(a)}~Time-lapse sequence ($t_1$--$t_5$) showing the drone
  (red box) descending toward and landing on a moving ground vehicle
  through approach, descent, and touchdown phases.
  \textbf{(b)}~3D trajectory overview: the UAV trajectory (blue) converges
  onto the vehicle trajectory (orange), with the ground projection
  (dashed) showing horizontal alignment.
  \textbf{(c)}~Altitude profile over time, illustrating smooth descent from
  12\,m to ground level.
  \textbf{(d)}~Horizontal convergence error, decreasing from
  ${\approx}6$\,m to within the $\pm 0.5$\,m tolerance band.
  All data are recorded within CARLA-Air's synchronous tick loop.%
}
\vspace{-2mm}
\label{fig:w1_coop}
\end{figure*}

\paragraph{Results.}
Fig.~\ref{fig:w1_coop} shows the complete landing sequence.
The drone descends from ${\approx}12$\,m to touchdown over ${\approx}20$\,s,
with horizontal convergence error decreasing from ${\approx}6$\,m to within
the $\pm 0.5$\,m tolerance band.
The 3D trajectory overview confirms that the UAV trajectory converges smoothly
onto the vehicle trajectory throughout the descent.
Table~\ref{tab:coop_results} summarizes the landing performance.


\begin{table}[!htbp]
\centering
\caption{%
  W1 cooperative precision landing results (Town10HD, RTX~A4000).%
}
\label{tab:coop_results}
\small
\setlength{\tabcolsep}{4pt}
\renewcommand{\arraystretch}{1.08}
\begin{tabular}{@{}lrl@{}}
\toprule
\textbf{Metric} & \textbf{Value} & \textbf{Notes} \\
\midrule
Mean FPS               & $19.3$             & Harmonic mean \\
Initial altitude       & ${\approx}12$\,m   & Start of descent \\
Landing duration       & ${\approx}20$\,s   & Approach to touchdown \\
Final horiz.\ error    & $< 0.5$\,m         & Within tolerance band \\
Initial horiz.\ error  & ${\approx}6$\,m    & At descent start \\
RPC errors             & $0$                & Both clients \\
\bottomrule
\end{tabular}
\end{table}

\subsection{W2: Embodied Navigation and VLN/VLA Data Generation}
\label{sec:apps:vln}

Vision-language navigation (VLN) and vision-language action (VLA) are among the
most active research directions in embodied intelligence, requiring agents to
navigate and act in realistic environments grounded in natural language
instructions and visual observations.
A key bottleneck is the availability of large-scale, diverse training data that
pairs language instructions with rich visual observations from multiple
viewpoints.
CARLA-Air provides a natural data generation infrastructure for this purpose:
its photorealistic urban environments, socially-aware pedestrians, dynamic
traffic, and simultaneous aerial-ground sensing enable the construction of
VLN/VLA datasets with cross-view visual grounding that single-domain platforms
cannot provide.

\paragraph{Platform capabilities for VLN/VLA.}
CARLA-Air supports VLN/VLA data generation through several platform-level
features:
(i)~both aerial and ground agents can be equipped with egocentric RGB, depth,
and semantic segmentation cameras, providing paired bird's-eye and street-level
visual observations along any navigation trajectory;
(ii)~the ground API exposes waypoint-based route planning with lane-level
precision, which can serve as the basis for generating language-grounded
navigation instructions;
(iii)~the shared rendering pipeline guarantees that all visual observations are
captured under identical weather, lighting, and scene conditions at each tick;
and (iv)~the aerial overview provides a natural ``oracle view'' for generating
spatial referring expressions and verifying navigation progress.

\begin{figure*}[!htbp]
\centering
\includegraphics[width=\textwidth]{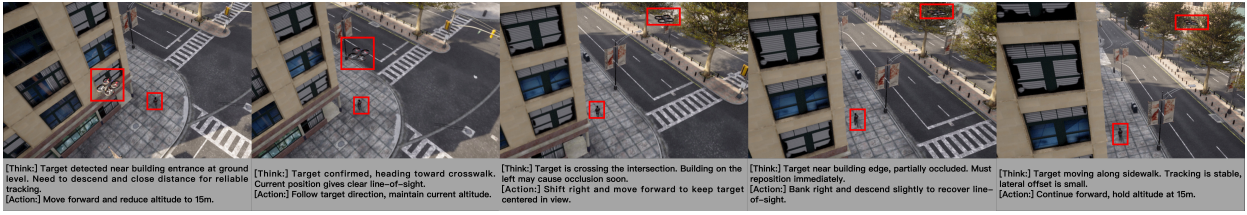}
\caption{%
  W2: Embodied navigation with aerial reasoning.
  A UAV autonomously tracks a pedestrian (red box, bottom) through an urban
  environment using bird's-eye visual observations.
  Each frame is annotated with the drone's chain-of-thought reasoning, illustrating how the agent interprets the scene,
  anticipates occlusions, and adjusts its flight path to maintain persistent
  visual contact with the target.
  All frames are rendered within CARLA-Air's shared simulation world under
  consistent lighting and physics.%
}
\label{fig:w2_vln}
\end{figure*}

\subsection{W3: Synchronized Multi-Modal Dataset Collection}
\label{sec:apps:dataset}

Generating large-scale paired aerial-ground datasets is a bottleneck for
training and evaluating cooperative perception models.
Manual synchronization across separate simulator processes introduces alignment
errors that corrupt correspondence annotations.
CARLA-Air eliminates this bottleneck: because both sensor suites are driven by
the same tick counter, the resulting dataset records carry guaranteed per-tick
correspondence with zero interpolation overhead.

\paragraph{Setup.}
Eight ground sensors (RGB, semantic segmentation, depth, LiDAR, radar, GNSS,
IMU, collision) and four aerial sensors (RGB, depth, IMU, GPS) are attached
concurrently.
The simulation runs in synchronous mode at a fixed timestep; all 12 sensor
callbacks are registered before the first tick advance.
Ground traffic is populated with 30 autopilot vehicles and 10 pedestrians.

\paragraph{Dataset structure.}
Each record $\mathcal{R}_k$ at tick $k$ contains all 12 sensor observations
sharing a common tick index, plus vehicle and drone pose in the unified world
frame.
Records are serialized to disk in a flat per-tick directory structure with
metadata in JSON.
No timestamp interpolation is required: the shared tick index serves as the
alignment key.

\paragraph{Results.}
Over 1\,000 ticks, the workflow produces 1\,000 fully synchronized 12-stream
records at a mean collection rate of ${\approx}17$\,Hz.
The maximum observed tick-alignment deviation across all 12 streams is one tick
(occurring transiently under disk-write saturation).
Table~\ref{tab:dataset_results} summarizes collection performance.


\begin{table}[!htbp]
\centering
\caption{%
  W3 multi-modal dataset collection results (1\,000 ticks, 30 vehicles,
  10 pedestrians, Town10HD, RTX~A4000).%
}
\label{tab:dataset_results}
\small
\setlength{\tabcolsep}{4pt}
\renewcommand{\arraystretch}{1.08}
\begin{tabular}{@{}lrl@{}}
\toprule
\textbf{Metric} & \textbf{Value} & \textbf{Notes} \\
\midrule
Mean FPS               & $17.1$         & Harmonic mean \\
Concurrent streams     & 12             & 8 ground + 4 aerial \\
Records collected      & 1\,000         & One per tick \\
Max alignment dev.     & ${\leq}1$ tick & Normal disk-write load \\
RPC errors             & $0$            & Both clients \\
Per-tick write latency & $61 \pm 9$\,ms & Incl.\ serialization \\
\bottomrule
\end{tabular}
\end{table}


\begin{figure*}[!htbp]
\centering
\includegraphics[width=\textwidth]{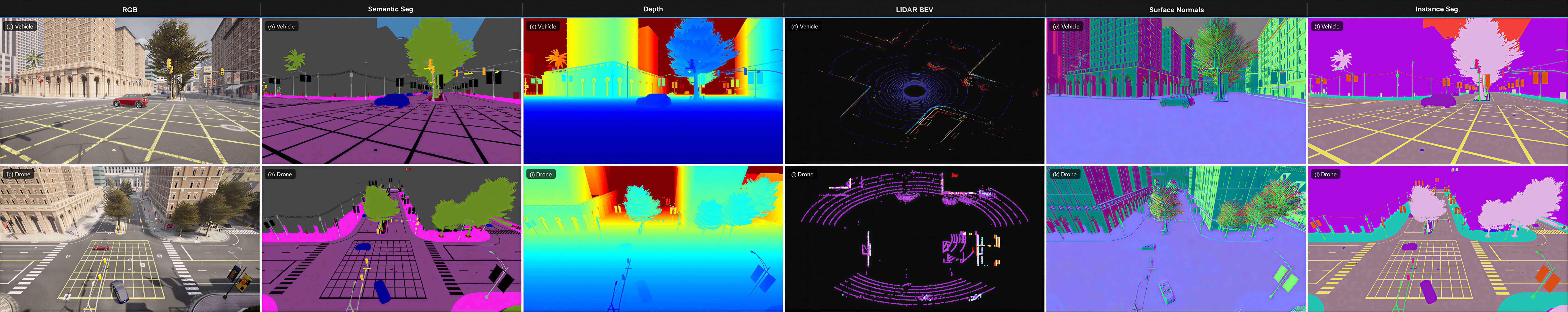}
\caption{%
  W3: Synchronized multi-modal dataset collection at a single simulation tick.
  \textbf{Top row} (vehicle perspective): RGB, semantic segmentation, depth,
  LiDAR bird's-eye view, surface normals, and instance segmentation.
  \textbf{Bottom row} (drone perspective): the same six modalities captured
  simultaneously from the aerial viewpoint.
  All 12 sensor streams share an identical tick index and are rendered under
  the same lighting and weather conditions through CARLA-Air's shared
  rendering pipeline, guaranteeing per-tick spatial-temporal correspondence
  without interpolation.%
}
\label{fig:w3_dataset}
\end{figure*}

\subsection{W4: Air-Ground Cross-View Perception}
\label{sec:apps:perception}

Cross-view perception---fusing aerial bird's-eye observations with ground-level
sensing---is an emerging research direction for cooperative autonomous driving,
urban scene understanding, and 3D reconstruction.
This workflow demonstrates CARLA-Air's ability to provide spatially and
temporally co-registered multi-modal sensor streams from both aerial and ground
viewpoints within a single shared environment.

\paragraph{Setup.}
A drone equipped with a depth camera hovers above a road segment while a ground
ego vehicle equipped with a semantic segmentation camera traverses the same
segment.
Both sensors are queried at each synchronous tick.

\paragraph{Sensor co-registration.}
Co-registration leverages the coordinate mapping from
Section~\ref{sec:arch:coords}.
The origin offset $\mathbf{d} \in \mathbb{R}^3$ between the aerial NED frame
and the ground world frame is computed once at initialization:
\begin{equation}
\mathbf{d} = T\!\left(\mathbf{p}_{\mathrm{spawn}}^{\mathrm{world}}\right)
  - \mathbf{p}_{\mathrm{spawn}}^{\mathrm{NED}},
\label{eq:coord_offset}
\end{equation}
where $T(\cdot)$ applies the scale conversion and axis remapping.
When the drone spawns at the world origin, $\mathbf{d} = \mathbf{0}$.

\paragraph{Weather consistency.}
To verify rendering consistency across both sensor layers, the workflow iterates
through all 14 official weather presets.
For each preset involving significant illumination change, the mean pixel
intensity of the aerial RGB frame shifts by more than 5\% relative to the
previous preset, confirming single-pass weather propagation.
Because both sensors share the same rendering pipeline, temporal alignment is
guaranteed:
\begin{equation}
\epsilon_k = \bigl|t_k^{\mathrm{gnd}} - t_k^{\mathrm{air}}\bigr| = 0,
\label{eq:sensor_alignment}
\end{equation}
a property that bridge-based architectures cannot provide.

\paragraph{Results.}
Over 500 ticks, the workflow produces 500 co-registered aerial-depth /
ground-segmentation pairs at ${\approx}18$\,Hz with zero RPC errors.
All 14 weather presets pass the illumination consistency assertion.
Table~\ref{tab:perception_results} summarizes the measured outcomes.


\begin{table}[!htbp]
\centering
\caption{%
  W4 cross-view perception results (500 ticks, Town10HD, RTX~A4000).%
}
\label{tab:perception_results}
\small
\setlength{\tabcolsep}{4pt}
\renewcommand{\arraystretch}{1.08}
\begin{tabular}{@{}lrl@{}}
\toprule
\textbf{Metric} & \textbf{Value} & \textbf{Notes} \\
\midrule
Mean FPS                & $18.2$            & Harmonic mean \\
Co-registered pairs     & 500               & Aerial depth + ground seg. \\
Per-tick latency        & $52 \pm 6$\,ms    & Full collection loop \\
Sensor alignment        & $0$ ticks         & Sync mode guarantee \\
Weather presets passed  & $14/14$           & All official presets \\
RPC errors              & $0$               & Both clients \\
\bottomrule
\end{tabular}
\end{table}

\begin{figure*}[!t]
\centering
\includegraphics[width=\textwidth, height=0.32\textheight, keepaspectratio]{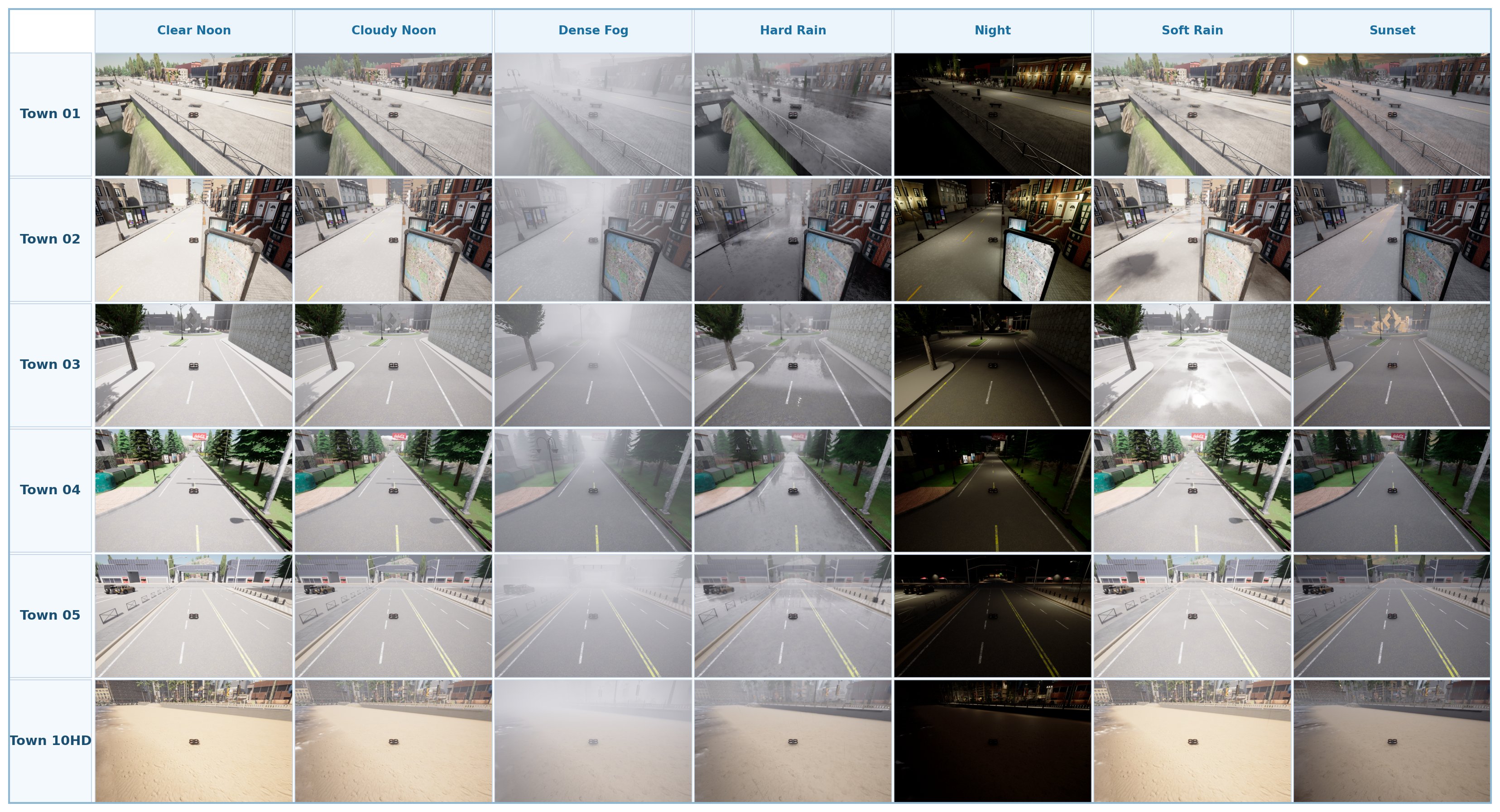}
\caption{%
  W4: Air-ground cross-view perception across diverse environments and weather
  conditions. Each row shows an aerial RGB view from the drone across six
  representative CARLA maps (Town\,01--05 and Town\,10HD); each column
  corresponds to a different weather preset (Clear Noon, Cloudy Noon, Dense
  Fog, Hard Rain, Night, Soft Rain, and Sunset).%
}
\label{fig:w4_perception}
\end{figure*}

\subsection{W5: Reinforcement Learning Training Environment}
\label{sec:apps:rl}

Reinforcement learning in air-ground cooperative settings requires a simulation
environment that provides closed-loop interaction, consistent state
observations across heterogeneous agents, and stable long-horizon episode
execution without memory leaks or synchronization drift.
CARLA-Air's single-process architecture naturally satisfies these requirements,
and the stability results from Section~\ref{sec:perf:vram} (zero crashes and
zero memory accumulation over 357 reset cycles) directly validate its
suitability as an RL training environment.

\paragraph{Platform capabilities for RL.}
CARLA-Air supports RL training workflows through several platform-level
features:
(i)~synchronous stepping mode provides deterministic state transitions
compatible with standard Gym-style training loops;
(ii)~both aerial and ground agents expose programmatic control interfaces (velocity
commands, waypoint targets, autopilot toggles) that can serve as action spaces;
(iii)~the full sensor suite (RGB, depth, segmentation, LiDAR, IMU, GPS)
provides rich observation spaces for both state-based and vision-based policies;
(iv)~episode resets via actor spawn/destroy are validated for stability across
hundreds of consecutive cycles (Section~\ref{sec:perf:vram});
and (v)~the shared world state ensures that reward signals computed from
cross-domain interactions (e.g., aerial-ground relative positioning) are
physically consistent.

\paragraph{Example: cooperative positioning.}
As a representative RL scenario, consider a drone learning to maintain an
optimal aerial observation position relative to a moving ground vehicle under
varying traffic conditions.
The observation space comprises the drone's pose, the vehicle's pose, and
surrounding traffic state; the action space is a 3D velocity command; the reward
encodes lateral tracking error and altitude maintenance.
This scenario exercises the full air-ground control loop within CARLA-Air's
synchronous tick, and can be implemented using standard RL libraries (e.g.,
Stable-Baselines3, RLlib) with minimal wrapper code.

\begin{figure*}[!htbp]
\centering
\includegraphics[width=\textwidth, height=0.35\textheight, keepaspectratio]{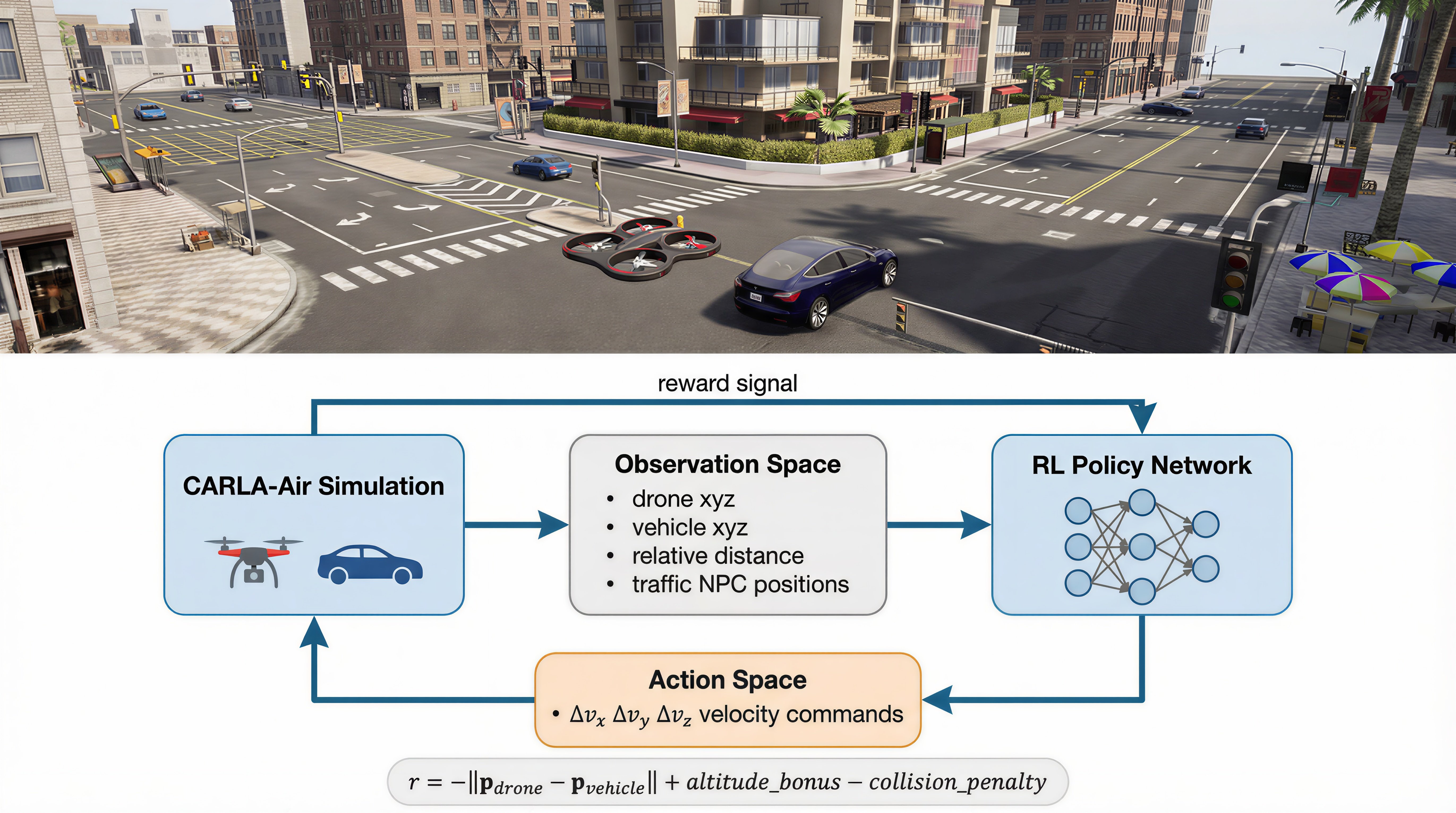}
\caption{%
  W5: Reinforcement learning training environment.
  \textbf{Top:} a drone learns to maintain an optimal aerial observation
  position above a moving ground vehicle within CARLA-Air's urban traffic
  environment.
  \textbf{Bottom:} the closed-loop RL pipeline. At each synchronous tick,
  CARLA-Air provides an observation space (drone and vehicle poses, relative
  distance, surrounding traffic state) to the policy network, which outputs
  3D velocity commands as actions. The reward signal encodes tracking
  accuracy, altitude maintenance, and collision avoidance.%
}
\label{fig:w5_rl}
\end{figure*}

\section{Limitations and Future Work}
\label{sec:limitations}

The current release of CARLA-Air is validated for the workflows presented in
Section~\ref{sec:apps}: single- and dual-drone aerial operations over
moderate-density urban traffic scenes.

\begin{itemize}[leftmargin=*]
  \item \textbf{Actor density.}
  Joint simulation performance is characterized at moderate traffic loads;
  high-density scenes with large simultaneous actor populations remain an
  active engineering target.

  \item \textbf{Environment resets.}
  Map switching requires a full process restart due to independent actor
  lifecycle management in each simulator backend; staged in-session resets are
  planned for a future release.

  \item \textbf{Multi-drone scale.}
  Configurations beyond two drones are functional but not yet formally
  validated across a wide range of scenarios; expanded multi-drone
  characterization will be documented once inter-agent behavior has been fully
  profiled.
\end{itemize}

\noindent None of these constraints affect the workflows in
Section~\ref{sec:apps}, all of which operate within the current boundaries.

Because CARLA-Air inherits and extends AirSim's aerial
capabilities---whose upstream development has been archived---long-term
maintenance of the aerial stack is managed within the CARLA-Air project
itself. Bug fixes, compatibility updates, and feature extensions to the aerial
subsystem are released as part of CARLA-Air's regular update cycle, ensuring
that the flight simulation capabilities continue to evolve independently of
the original upstream repository.

Looking ahead, near-term work will address physics-state synchronization
between the two engines and a ROS\,2~\cite{ros2} bridge that republishes both
simulator streams as standard topics for broader ecosystem integration.
Longer-term, we aim to support GPU-parallel multi-environment execution in the
spirit of Isaac Lab~\cite{isaaclab} and OmniDrones~\cite{omnidrones}, bringing
CARLA-Air closer to the episode throughput required for large-scale
reinforcement learning.

\section{Conclusion}
\label{sec:conclusion}

Simulation platforms for autonomous systems have historically fragmented along
domain boundaries, forcing researchers whose work spans ground and aerial
domains to maintain inter-process bridge infrastructure or accept capability
compromises.
CARLA-Air resolves this fragmentation by integrating CARLA~\cite{carla} and
AirSim~\cite{airsim} within a single Unreal Engine process~\cite{ue4},
exposing both native Python APIs concurrently over a shared physics tick and
rendering pipeline.

\paragraph{Technical contributions.}
The central technical contribution is a principled resolution of the
single-GameMode constraint through a composition-based design:
\texttt{CARLAAirGameMode} inherits CARLA's ground simulation subsystems while
composing AirSim's aerial flight actor as a standard world entity, with
modifications to exactly two upstream source files.
This design yields three properties unavailable in bridge-based approaches: a
shared physics tick that eliminates inter-process clock drift, a shared
rendering pipeline that guarantees consistent weather and lighting across all
sensor viewpoints, and full preservation of both upstream Python APIs.

\paragraph{Platform capabilities.}
Building on this architecture, CARLA-Air provides a photorealistic, physically
coherent simulation world with rule-compliant traffic, socially-aware
pedestrians, and aerodynamically consistent multirotor dynamics.
Up to 18 sensor modalities can be synchronously captured across aerial and
ground platforms at each simulation tick.
The platform directly supports four research directions in air-ground embodied
intelligence: air-ground cooperation, embodied navigation and vision-language
action, multi-modal perception and dataset construction, and
reinforcement-learning-based policy training.

\paragraph{Validation.}
The platform is validated through performance benchmarks demonstrating stable
operation at ${\approx}20$\,FPS under joint workloads, a 3-hour continuous
stability run with zero crashes across 357 reset cycles, and five
representative workflows that exercise the platform's core capabilities.
A consolidated feature comparison with representative platforms is provided in
Table~\ref{tab:platform_comparison} (Section~\ref{sec:related}).

\paragraph{Broader impact.}
By providing a shared world state for aerial and ground agents, CARLA-Air
enables research directions that are structurally inaccessible in single-domain
platforms: paired aerial-ground perception datasets from physically consistent
viewpoints, coordination policies over joint multi-modal observation spaces,
and embodied navigation grounded in cross-view visual and linguistic input.
By also inheriting and extending the aerial simulation capabilities of
AirSim---whose upstream development has been archived---CARLA-Air ensures that
this widely adopted flight stack continues to evolve within a modern, actively
maintained infrastructure.

\bibliographystyle{plain}
\bibliography{references}

\appendix

\appendix
\section{Appendix}
\subsection{System Configuration}
\label{app:config}

\begin{figure}[!htbp]
\centering
\includegraphics[width=\columnwidth, height=0.3\textheight, keepaspectratio]{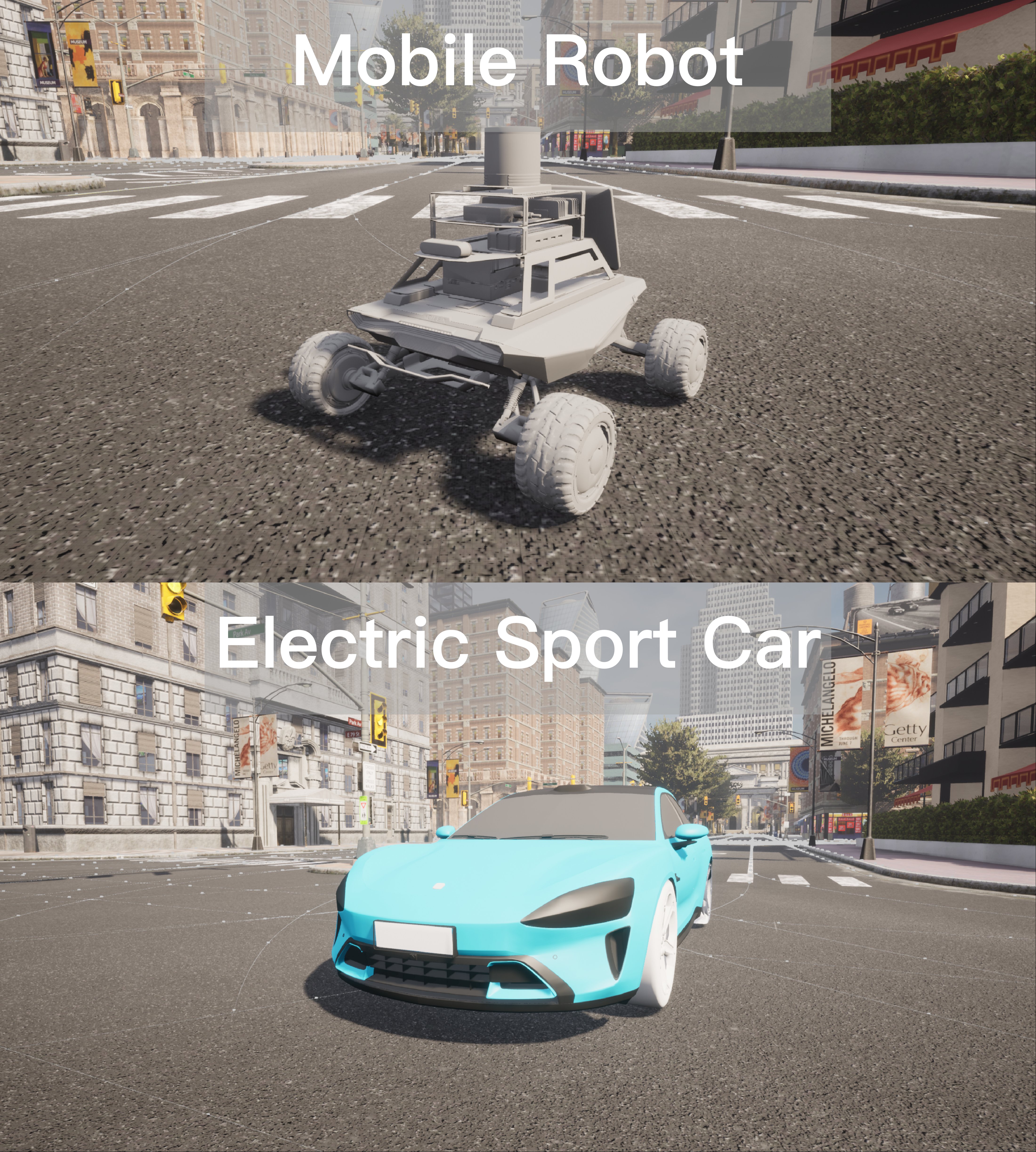}
\caption{%
  Custom assets imported into CARLA-Air through the extensible asset pipeline.
  \textbf{Top:} a four-wheeled mobile robot with onboard LiDAR, imported from
  an external FBX model.
  \textbf{Bottom:} a custom electric sport car with user-defined vehicle
  dynamics.
  Both assets operate within the shared simulation world alongside all
  built-in CARLA traffic and AirSim aerial agents, and are visible to all
  sensor modalities.%
}
\label{fig:custom_assets}
\end{figure}

\paragraph{Reference hardware.}
All experiments in this report were conducted on the following configuration:
Ubuntu 20.04/22.04~LTS, NVIDIA RTX~A4000 (16\,GB GDDR6), AMD Ryzen~7 5800X
(8-core, 4.7\,GHz), 32\,GB DDR4-3200.

\paragraph{Software stack.}
CARLA~0.9.16, AirSim~1.8.1 (final stable open-source release),
Unreal Engine~4.26, Python~3.8+.

\paragraph{Network configuration.}
Table~\ref{tab:ports} lists the default port assignments.
Both RPC servers bind to \texttt{localhost} by default; remote connections
require explicit IP configuration.

\begin{table}[!htbp]
\centering
\caption{Default network port assignments.}
\label{tab:ports}
\footnotesize
\setlength{\tabcolsep}{4pt}
\renewcommand{\arraystretch}{1.08}
\begin{tabular}{@{}llr@{}}
\toprule
\textbf{Service} & \textbf{Protocol} & \textbf{Port} \\
\midrule
CARLA RPC Server   & TCP & 2000  \\
CARLA Streaming    & UDP & 2001  \\
AirSim RPC Server  & TCP & 41451 \\
\bottomrule
\end{tabular}
\end{table}

\paragraph{Distribution.}
The prebuilt binary package is approximately 19\,GB and includes a one-command
launcher (\texttt{CarlaAir.sh}).
The source distribution is approximately 651\,MB and is released under the MIT
license.

\subsubsection{API Compatibility Summary}
\label{app:api}

Table~\ref{tab:api_compat} summarizes the API compatibility status and test
coverage of the current CARLA-Air release.
All 89 automated CARLA API tests pass without modification; the full AirSim
flight control and sensor access API has been verified through manual and
scripted testing.
A total of 63 ROS\,2 topics are published across both simulation backends.

\begin{table}[!htbp]
\centering
\caption{API compatibility and test coverage.}
\label{tab:api_compat}
\small
\setlength{\tabcolsep}{3pt}
\renewcommand{\arraystretch}{1.08}
\resizebox{\columnwidth}{!}{%
\begin{tabular}{@{}lp{4.5cm}@{}}
\toprule
\textbf{Component} & \textbf{Status} \\
\midrule
CARLA API           & 89/89 automated tests passing \\
AirSim API          & Full flight control and sensor access verified \\
ROS\,2 topics       & 63 total (43 CARLA + 14 AirSim + 6 generic) \\
\midrule
\multicolumn{2}{@{}l}{\textit{Key upstream scripts confirmed working}} \\
\texttt{manual\_control.py}    & CARLA manual driving interface \\
\texttt{automatic\_control.py} & CARLA autopilot demonstration \\
\texttt{dynamic\_weather.py}   & CARLA weather preset cycling \\
\texttt{hello\_drone.py}       & AirSim basic flight demonstration \\
\bottomrule
\end{tabular}}%
\end{table}

\subsection{Upstream Source Modifications}
\label{app:source_mods}

CARLA-Air is designed to minimize modifications to the upstream CARLA codebase.
The integration touches only two header files and one source file, totaling
approximately 35 lines of changes.
All other integration code is purely additive, residing within the aerial
simulation plugin as the \texttt{CARLAAirGameMode} class
(${\sim}1{,}405$ lines of C++).
Table~\ref{tab:source_mods} provides the complete modification summary.

\begin{table}[!htbp]
\centering
\caption{Upstream CARLA source modifications.}
\label{tab:source_mods}
\small
\setlength{\tabcolsep}{3pt}
\renewcommand{\arraystretch}{1.08}
\resizebox{\columnwidth}{!}{%
\begin{tabular}{@{}lp{4.8cm}@{}}
\toprule
\textbf{File} & \textbf{Modification} \\
\midrule
\texttt{CarlaGameModeBase.h}
  & \texttt{friend} declaration for \texttt{CARLAAirGameMode} \\
\texttt{CarlaEpisode.h}
  & Sensor tagging visibility: \texttt{private} $\to$ \texttt{protected} \\
\texttt{CarlaGameModeBase.cpp}
  & State flag assignment (1 line) \\
\midrule
\multicolumn{2}{@{}l}{\textit{New integration layer (no upstream modification)}} \\
\texttt{CARLAAirGameMode}
  & Unified game mode class; ${\sim}1{,}405$ lines C++ \\
\bottomrule
\end{tabular}}%
\end{table}

\subsection{Custom Asset Import}
\label{app:assets}

CARLA-Air supports importing custom 3D assets (robot platforms, vehicles, UAV
models, and environment objects) through an asset pipeline built on Unreal
Engine's content framework. Imported assets are registered as spawnable actor
classes and become accessible through the standard CARLA Python API
(\texttt{world.spawn\_actor()}). Once registered, custom assets share the same
physics tick, rendering pass, and sensor visibility as all built-in actors,
ensuring full consistency within the joint simulation environment.

\end{document}